\pdfoutput=1

\documentclass{article}
\usepackage[letterpaper,top=2cm,bottom=2cm,left=3cm,right=3cm,marginparwidth=1.75cm]{geometry}

\usepackage{algorithm}
\usepackage[noend]{algpseudocode}
\usepackage{amsfonts}
\usepackage{amsmath}
\usepackage{booktabs}
\usepackage{caption}
\usepackage{hyperref}
\usepackage{makecell}
\usepackage{multirow}
\usepackage{natbib}
\usepackage{wrapfig}
\usepackage{graphicx}
\usepackage{subcaption}
\usepackage{xcolor}

\usepackage{times}
\usepackage{latexsym}

\usepackage[T1]{fontenc}

\usepackage[utf8]{inputenc}

\usepackage{microtype}

\usepackage{inconsolata}

\newcommand{\cA}{\mathcal{A}}

\newcommand{\cU}{\mathcal{U}}

\newcommand{\bW}{\mathbf{W}}

\newcommand{\cN}{\mathcal{N}}

\newcommand{\cE}{\mathcal{E}}

\newcommand{\bt}{\mathbf{t}}

\newcommand{\bp}{\mathbf{p}}

\newcommand{\bS}{\mathbf{S}}

\newcommand{\bw}{\mathbf{w}}

\newcommand{\cF}{\mathcal{F}}
\newcommand{\cM}{\mathcal{M}}

\newcommand{\plum}{{\texttt{Plum}}}
\newcommand{\plumhc}{{\texttt{Plum-HC}}}
\newcommand{\plumsa}{{\texttt{Plum-SA}}}
\newcommand{\plumgam}{{\texttt{Plum-GA-M}}}
\newcommand{\plumgac}{{\texttt{Plum-GA-C}}}
\newcommand{\plumts}{{\texttt{Plum-TS}}}
\newcommand{\plumhs}{{\texttt{Plum-HS}}}

\title{Plum: Prompt Learning using Metaheuristic}

\author{
\bf Rui Pan$^1$\thanks{\quad Equal contribution.},\quad
Shuo Xing$^2$\footnotemark[1],\quad
Shizhe Diao$^1$\footnotemark[1],\quad
Wenhe Sun$^1$\footnotemark[1] \thanks{\quad Work done during internship at the Hong Kong University of Science and Technology.},\quad
Xiang Liu$^1$\footnotemark[2],\quad \\
\bf Kashun Shum$^1$,\quad
Jipeng Zhang$^1$,\quad
Renjie Pi$^1$,\quad
Tong Zhang$^3$\\
\\
$^1$The Hong Kong University of Science and Technology \quad
$^2$Texas A\&M University \\
$^3$University of Illinois Urbana-Champaign \\
\texttt{\{rpan,sdiaoaa,ksshumab,jzhanggr,rpi\}@ust.hk,}\quad \texttt{shuoxing@tamu.edu,} \\
\texttt{sunwenhe@mail.nankai.edu.cn},\quad  \texttt{xliu886@connect.hkust-gz.edu.cn},\\
\texttt{tongzhang@tongzhang-ml.org}
}

\begin{document}
\maketitle
\begin{abstract}
\noindent Since the emergence of large language models, prompt learning has become a popular method for optimizing and customizing these models. 
Special prompts, such as Chain-of-Thought, have even revealed previously unknown reasoning capabilities within these models. 
However, the progress of discovering effective prompts has been slow, driving a desire for general prompt optimization methods. 
Unfortunately, few existing prompt learning methods satisfy the criteria of being truly "general", i.e., automatic, discrete, black-box, gradient-free, and interpretable all at once. 
In this paper, we introduce metaheuristics, a branch of discrete non-convex optimization methods with over 100 options, as a promising approach to prompt learning.
Within our paradigm, we test six typical methods: hill climbing, simulated annealing, genetic algorithms with/without crossover, tabu search, and harmony search, demonstrating their effectiveness in white-box and black-box prompt learning. Furthermore, we show that these methods can be used to discover more human-understandable prompts that were previously unknown in both reasoning and image generation tasks, opening the door to a cornucopia of possibilities in prompt optimization. We release all the codes in \url{https://github.com/research4pan/Plum}.
\end{abstract}

\section{Introduction}

The advent of powerful large language models (LLMs)~\citep{devlin2018bert, radford2019gpt2,brown2020gpt3,team2023gemini,roziere2023codellama,touvron2023llama,touvron2023llama2} has paved the way for numerous real-world applications, including multi-round chat~\citep{openai2023gpt4,touvron2023llama2}, instruction following~\citep{ouyang2022instructgpt} and reasoning~\citep{roziere2023codellama,azerbayev2023llemma}. To further enhance the performance of LLMs in specific domains, additional optimization techniques are commonly employed, such as finetuning~\citep{chiangvicuna}, adapter methods~\citep{hu2021lora,dettmers2023qlora}, alignment approaches~\citep{ouyang2022instructgpt,dong2023raft}, and prompt learning~\citep{sun2022black,diao2022black}. 
Notably, prompt learning distinguishes itself from other methods by eliminating the need for gradient information from models, resulting in substantially reduced memory consumption and computational resource requirements. Furthermore, prompt learning often yields interpretable outcomes, which help researchers and engineers intuitively understand its effectiveness, thereby being beneficial in inspiring more generalizable prompts for various tasks~\citep{prasad2022grips,guo2023evoprompt,yu2023gptfuzzer}.

Since the introduction of prompt engineering and prompt learning, significant advancements have been made in the discovery of effective prompts. A noteworthy illustration is Chain-of-Thought (COT), whereby the simple inclusion of accurate deduction steps for few-shot examples within the original prompt empowers LLMs to achieve substantial performance improvements in reasoning tasks~\citep{wei2022cot}. 
An even more impressive result is Zero-shot-COT~\cite{kojima2022zerocot}, where adding the magic phrase ``Let's think step by step'' produces a remarkable accuracy gain of over 10\% across multiple models engaged in a diverse spectrum of reasoning tasks.

However, the quest for such highly effective prompts remains unfulfilled, underscoring the need for tools that accelerate the discovery process. Ideally, these prompt learning tools should possess a notable level of generality, while simultaneously meeting the following criteria:
\begin{itemize}
  \item Automatic: since human involvements are normally expensive and time-consuming.
  \item Discrete: as commercial LLMs usually provide APIs that only accept discrete inputs~\citep{openai2023gpt4,team2023gemini}.
  \item Black-box: applicable to black-box LLMs.
  \item Gradient-free: hence still possess the good property of low memory cost and time cost.
  \item Interpretable: thus researchers can understand the logics behind its effectiveness and extend its applications with ease.
\end{itemize}

Regretfully, to the best of our knowledge, only a limited number of prompt learning methods meet the criteria for being considered ``general'' as per the aforementioned definition. Manual prompt engineering techniques employing handcrafted prompts~\citep{mishra2021reframing, kojima2022zerocot,bsharat2023principled} lack automation, while continuous prompt learning approaches~\citep{liu2021gpt,li2021prefix} fail to satisfy the requirements of discreteness and interpretability. Prompt learning methods employing reinforcement learning~\citep{deng2022rlprompt} often necessitate an additional neural network, the learning process of which is typically non-interpretable, and their theoretical guarantees are only attainable under stringent assumptions.

In this paper, we present a novel approach to simultaneously achieve these desirable properties by integrating prompt learning with metaheuristics.
By treating prompt learning as a non-convex discrete optimization problem within a black-box framework, we harness the potential of metaheuristics, which offer interpretable and automated optimization processes.
Moreover, this domain of optimization research encompasses a vast array of over 100 distinct algorithms~\citep{hussain2019metaheuristic}, ushering in a new paradigm for prompt learning.
Many of these algorithms even boast robust theoretical guarantees, ensuring the discovery of both local and even global optimal solutions within limited time frames~\citep{locatelli2000satheory,glover2002tabu,schmitt2004gatheory,dorigo2006ant,he2018cuckootheory}.

\section{Related Work}

\subsection{Prompt Learning}

Prompt-based learning is an effective approach that harnesses the power of large language models (LLMs) to facilitate downstream tasks by extracting relevant knowledge. This approach offers several advantages, notably in terms of computational efficiency, as only a small set of parameters needs to be optimized, in contrast to traditional methods that require fine-tuning the entire LLM for each task.
Prompting methods can be categorized into discrete prompts~\citep{davison2019commonsense, wallace2019universal, jiang2020can, shin2020autoprompt, haviv2021bertese, yuan2021bartscore, gao2020making, ben2021pada, su2022selective, diao2022black, datta2023prompt, pan2024pomp} and continuous prompts~\citep{hambardzumyan2021warp, zhong2021factual, han2021ptr,li2021prefix, qin2021learning, liu2021gpt} based on the format of prompts. 
Discrete prompts are typically represented as token sequences or natural language phrases, while continuous prompts are designed as sequences of vectors.
Significant advancements have been made in leveraging LLMs for reasoning tasks using chain-of-thought prompting techniques. Notable progress is evidenced by the research of ~\citet{wei2022chain, wang2022self, zhou2023leasttomost, zhang2022automatic, shum2023automatic, diao2023active}, in which well-structured prompts guide the models to showcase superior reasoning capabilities, surpassing previous benchmarks. Further studies in prompt attacks~\citep{yu2023gptfuzzer,zhu2023promptbench} have also utilized several of these advancements to improve the robustness of modern LLMs.
However, most of the prior studies on prompting have been limited to a white-box setting, requiring access to all parameters of a large language model. The reliance on white-box optimization methods has become less practical in light of the widespread use of closed-source black-box models.

\subsection{Black-Box Prompt Learning}
Black-box prompt learning~(BPL)~\citep{diao2022black, sun2022black, sun2022bbtv2} represents a promising research direction that tunes prompts without requiring access to the parameters and gradients of the underlying LLMs. BPL proves particularly valuable for closed-source models, which have demonstrated superior performance compared to open-source models~\citep{liang2022holistic}. This advancement allows for the effective optimization of closed-source models, overcoming the previous limitations posed by the unavailability of model parameters.
Similar to white-box prompts, black-box prompts can be categorized into discrete prompts~\citep{diao2022black, prasad2022grips, deng2022rlprompt, hou2022promptboosting, cheng2023black} and continuous prompts~\citep{sun2022black, sun2022bbtv2, su2022few}. However, in the black-box setting, discrete prompts prove more practical as the interface of LLMs only accepts discrete inputs and cannot process continuous vectors.

\subsection{Metaheuristics}
There have been efforts combining the power of metaheuristics with black-box prompt learning in the past literature. 
For instance, GrIPS~\citep{prasad2022grips} applied greedy search methods, as well as simulated annealing~\citep{kirkpatrick1983optimization}, to search better prompts with simple edit operations such as addition, swap, paraphrase, and deletion. 
GPS~\citep{xu2022gps} utilized genetic algorithms for few-shot prompt learning, improving handcrafted prompts on multiple tasks. 
Additionally, \citet{kumar2021reordering} and \citet{lu2021fantastically} investigated the impact of example orders in few-shot learning settings, respectively improving performance through the use of genetic algorithms and alternate search heuristics. However, these methods are limited in their applicability, focusing on specific prompt learning settings and failing to fully explore the inherent potential of discrete optimization, which is a bridge between black-box prompt learning and general metaheuristics. As an attempt to overcome this limitation, \citep{guo2023evoprompt} investigated the performance of combining the rephrasing power of LLMs with Genetic Algorithm and Differential Evolution, showing superior performance over hand-crafted prompts. However, compared with our work, \citet{guo2023evoprompt} focused more on leveraging the paraphrasing ability of LLMs and restricted search strategy to only evolutionary algorithms, which is a subset of metaheuristics.
 
Metaheuristics is a well-established and versatile branch of discrete optimization, known for its effectiveness in solving various non-convex optimization problems. This field has witnessed the development of over 100 different methods~\citep{hussain2019metaheuristic}, which have been successfully applied to a wide range of problem domains. For instance, they have been utilized in Neural Architecture Search~\citep{elsken2019nas,liu2021evonas}, Travelling Salesman Problem~\citep{ouaarab2014tspcuckoo,hussain2017tspga,grabusts2019tspsa}, and Program Search~\citep{chen2023lion}. Prominent algorithms in this field include simulated annealing~\citep{kirkpatrick1983optimization}, genetic algorithms~\citep{holland1992genetic}, tabu search~\citep{glover1986tabu}, harmony search~\citep{geem2001harmonysearch}, ant colony optimization~\citep{dorigo1997ant}, among others. However, the integration of these powerful metaheuristics with black-box prompt learning remains largely unexplored, presenting rich opportunities for further research in this direction.

\section{Methods}
In this paper, we propose Prompt learning using metaheuristic (\plum{}), a general paradigm that enables the application of numerous metaheuristics in the setting of discrete black-box prompt learning. 
Based on this framework, the flexible composition of different techniques becomes possible, leading to a ``template'' for designing general prompt learning algorithms, guaranteed to simultaneously satisfy the properties of automatic, discrete, black-box, gradient-free, and interpretable.

\subsection{Plum in General}
\begin{algorithm}
\caption{\plum{} in General}
\label{alg:plum_general}
\textbf{Required:}
\\ (1) A well-defined neighbor set $\cN(\bp)$ for $\forall \bp \in \Omega$;
\\ (2) a metaheuristics algorithm $\cA$;
\\ (3) metaheuristics-dependent hyperparameters $\theta$ and functions 
$\cF$.
\\
\\
\textbf{Input:} An initialized prompt $\bp_0$ and an objective $f(\bp)$ to be maximized (or minimized).
\begin{algorithmic}[1]
  \State $\bp_\ast \gets$ \Call{$\cA_{\cN, \theta, \cF, f}$}{$\bp_0, \cM$}
  \State \Return $\bp_\ast$
\end{algorithmic}
\end{algorithm}

The framework of \plum{} is outlined in Algorithm~\ref{alg:plum_general}, which encompasses four fundamental and orthogonal elements. The first element is a well-defined neighborhood $\cN(\cdot)$ in the discrete prompt search space. In practice, any operations that transform a prompt into another prompt can be used to define such neighborhoods. For instance, edit operations in GrIPS~\citep{prasad2022grips} can convert a prompt $\bp = [ \text{Let}, \text{us}, \text{think} ]$ to a new prompt $\bp' = [ \text{Let}, \text{us}, \text{brainstorm} ]$ via paraphrase, and all such possible conversions form prompt $\bp$'s neighborhood
\begin{align*}
  \cN(\bp) = \{ \bp' | \exists e \in \cE, s.t. \ \bp' = \texttt{edit}(\bp, e) \}.
\end{align*}
This abstraction of neighborhood decouples prompt transformations and search algorithms, making the flexible combination of those techniques possible.

The second part is the core metaheuristics $\cA$, such as simulated annealing~\citep{kirkpatrick1983optimization}, tabu search~\citep{glover1986tabu}, genetic algorithms~\citep{holland1992genetic}. Any discrete optimization metaheuristics can fit into this part.

Notably, various metaheuristics come with their own specific hyperparameters and functions that need to be considered within the \plum{} paradigm. For instance, in the case of simulated annealing, a temperature scheduler is required for regulating the annealing speed, where an appropriate schedule can greatly improve the search efficiency. Similarly, general genetic algorithms incorporate a ``crossover'' operation, which swaps segments of two prompts and serves as a global search mechanism to expand its search scope.

\begin{table*}[t]
  \footnotesize
  \begin{center}
    \begin{tabular}{ccccl}
      \toprule

      \makecell{\textbf{Neighborhood}\\ $\cN(\bp)$}
      & \makecell{\textbf{Metaheuristics}\\$\cA$}
      & \makecell{\textbf{Hyperparameter}\\$\theta$}
      & $\cF$
      & Name
      \\
      \midrule

      \multirow{15}{*}{\makecell{Candidates \\after editing\\
      \citep{prasad2022grips}}}
      & Hill Climbing
      & -
      & -
      & \plumhc{}
      \\
      \cmidrule(l){2-5}

      & Simulated Annealing
      & -
      & \makecell{Temperature\\ schedule $T(i)$}
      & \plumsa{}
      \\
      \cmidrule(l){2-5}

      & \makecell{Genetic Algorithms \\(mutation only)}
      & \makecell[l]{$k$ tournament selections}
      & -
      & \plumgam{}
      \\
      \cmidrule(l){2-5}

      & \makecell{Genetic Algorithms \\(with crossover)}
      & \makecell[l]{1) $k$ tournament selections;\\
      2) $p_{\text{mutation}}$: mutation rate}
      & \makecell{Crossover function: \\$\langle \bp_1, \bp_2 \rangle \rightarrow \bp$}
      & \plumgac{}
      \\
      \cmidrule(l){2-5}

      & \makecell{Tabu Search}
      & \makecell[l]{$N_{\text{tabu}}$ slots in Tabu list}
      & \makecell{Tabu function:\\$\langle\mathcal{T}, \bp\rangle \to \{0,1\}$}
      & \plumts{}
      \\
      \cmidrule(l){2-5}

      & \makecell{Harmony Search}
      & \makecell[l]{1) $N_H$ harmony search memory;\\
      2) $k_s$ number of segments;\\
      3) $HMCR \in [0, 1]$: harmony\\ memory considering rate;\\
      4) $PAR \in [0, 1]$: pitching adjust \\rate}
      & -
      & \plumhs{}
      \\

      \bottomrule
    \end{tabular}
  \end{center}
  \caption{Implemented \plum{} algorithms. }
  \label{tab:plum_impl}
\end{table*}

During the optimization process, the objective is to maximize (or minimize) a provided function $f(\bp)$, which typically corresponds to task-specific performance metrics of the target LLM. For instance, under the setting of GPT-3-babbage~\citep{brown2020gpt3} on Commonsense Question Answering~\citep{talmor-etal-2019-commonsenseqa}, the function $f(\bp)$ represents the final accuracy achieved by augmenting GPT-3-babbage with the prompt $\bp$. The primary goal of the metaheuristics is to search for the optimum $\bp$ which maximizes this accuracy.

By clearly stating those key components in \plum{}, we are now prepared to instantiate \plum{} with specific metaheuristics and neighborhood definitions that can perform real-world prompt learning tasks.

\subsection{Plum in Practice}

As a proof of concept, we realize \plum{} with several popular metaheuristics, as listed in Table~\ref{tab:plum_impl}. Detailed pseudocodes of those methods are available in Appendix~\ref{appendix:alg-details}.

All those algorithms are successful products of metaheuristics under the setting of prompt learning. 
By combining the edit operations employed in GrIPS~\citep{prasad2022grips} with simulated annealing~\citep{kirkpatrick1983optimization}, we developed an algorithm capable of recovering global optimum with proper temperature scheduler ~\citep{granville1994saconverge}. 
In this context, we define the global optimum as the prompt that maximizes the objective function $f(\bp)$ within the set of reachable prompts starting from the initial prompt $\bp_0$.

Furthermore, replacing simulated annealing with genetic algorithms gives birth to \plumgam{} and \plumgac{}, whose metaheuristics prototypes are proven to converge to global optimum given specific mutation and crossover operations~\citep{schmitt2004gatheory}.
Compared with \plumgam{}, \plumgac{} adopts an extra operation of crossover, which enables more aggressive searches.
Intuitively, \plumgac{} > \plumgam{} > \plumsa{} > \plumhc{} in terms of their exploration power, which is at the expense of increased search steps and the number of evaluations of the objective function $f(\bp)$.

Similarly, adopting Tabu Search~\citep{glover1986tabu} and Harmony Search~\citep{geem2001harmonysearch} brings forth \plumts{} and \plumhs{}, both are API-efficient methods according to our experiments. Here Tabu Search introduces a Tabu list to prohibit certain candidates from being revisited, hence avoids the search process from getting stuck in a local optimum. Meanwhile, Harmony Search obtains its inspiration from musical composition, utilizing a harmony search memory with past prompts and generates the new prompt by combining segments from them.

\begin{figure*}[t]
  \centering
   \includegraphics[scale=0.55]{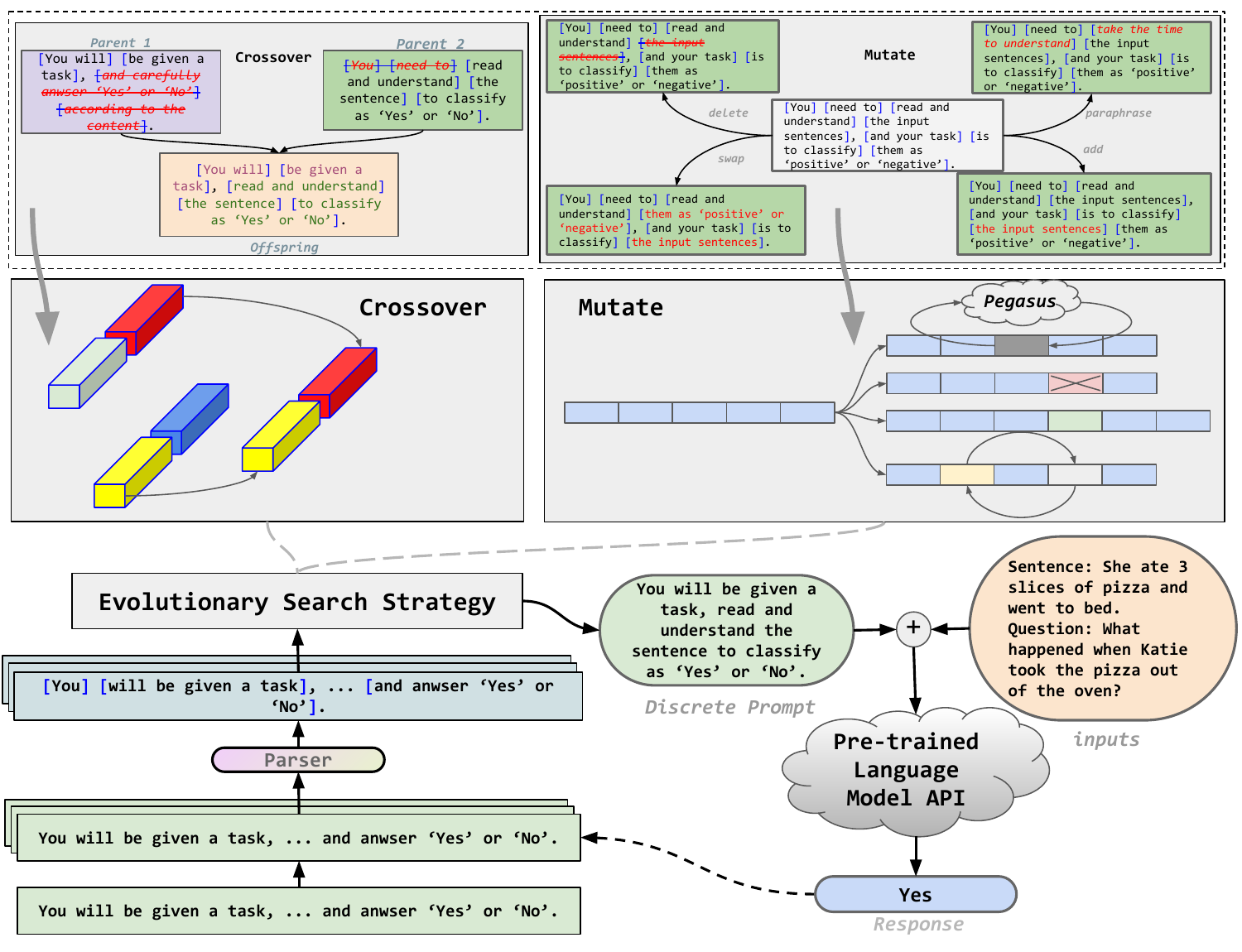}
  \caption{Illustration of \plumgac{} and its possible extension with LLMs.}
  \label{fig:pipeline}
\end{figure*}

\subsection{Plum-GA-C}\label{sec:plum-ga-c-details}
Here we illustrate one specific algorithm \plumgac{} as an example of instantiating a \plum{} algorithm. \plumgac{} contains three stages: parsing, editing, and scoring, which are described as follows.

\subsubsection{Stage 1: Parsing}
\label{sec:plum-ga-c-parsing}
Initially, we parse the initial prompt $I$ into a sequence of phrases using a CRF-based constituency parser~\citep{zhang2020fast}.
Subsequently, we iteratively merge the leaf nodes of the obtained constituency tree until several disjoint phrase-level constituents are generated.

\subsubsection{Stage 2: Editing}
\label{sec:plum-ga-c-editing}
\paragraph{Mutation} Mutation is a combination of edit operations inside the prompt. 
Following~\citet{prasad2022grips}, we introduce four mutation operations including: \texttt{delete}, \texttt{add}, \texttt{swap}, and \texttt{paraphrase}.
\texttt{delete} randomly drops some phrases while \texttt{addition} adds some randomly sampled phrases back at a random position.
\texttt{swap} swaps two phrases, and \texttt{paraphrase} replaces a phrase with a new phrase generated by a paraphrasing model. 
We apply PEGASUS~\citep{zhang2020pegasus} as the paraphrase model to obtain new phrases.
The upper right corner of Figure~\ref{fig:pipeline} provides an illustrative depiction of the mutation process. 

\paragraph{Crossover} Crossover is a combination of edit operations across two prompts.
To perform crossover, we divide each prompt into two sub-sequences at a randomly selected position. From each prompt, we randomly choose one sub-sequence and these selected sub-sequences are then combined to create a new sequence as the prompt. The upper left corner of Figure~\ref{fig:pipeline} provides a visual representation of the crossover process.

\subsubsection{Stage 3: Scoring}
\label{sec:plum-ga-c-scoring}
The optimization process relies on performance signals acquired through the scoring process. Initially, we construct a score set $S = \{(x_1, y_1), (x_2, y_2), $ $..., (x_n, y_n)\}$ by randomly sampling $n$ examples from the training split. Subsequently, a score function is computed based on the accuracy of predictions.

\subsection{Plum-HS}
\label{sec:plum-hs}
We also introduce \plumhs{}, a novel algorithm that integrates Harmony Search into prompt learning. Notably, it marks the first instance of Harmony Search being successfully applied to prompt optimization tasks. The process is quite similar to composing new songs, where new prompts are generated by combining segments from past prompts together. The procedure starts with a base prompt pending to be optimized, which is first divided into segments of words, and inserted into the harmony memory.

At each iteration, a new prompt is generated by composing $k_s$ segments from prompts in the harmony search memory, where the $j$-th segment is the exact copy of the $j$-th segment from a randomly sampled prompt in the memory. After this composition procedure, a pitch finetuning process is then applied, which edits the segment with a small probability in a similar manner as the mutation operation in Section~\ref{sec:plum-ga-c-editing}. The only difference is that it utilizes two kinds of pitch change operations: small pitch changes which only do paraphrasing, and big pitch changes which include all editing operators. The generated new prompt is then evaluated and inserted back into the harmony search memory, where the memory only retains $N_H$ prompts with top scores.

Intuitively speaking, \plumhs{} has a similar exploration power as \plumgac{} and adopts a more flexible manner to replace long segments in the middle of the prompt. According to our experiments, it is also API-efficient, which achieves better performance while consuming significantly fewer API calls compared to other methods. 

Besides \plumgac{} and \plumhs{}, details of other algorithms are also available in Appendix~\ref{appendix:alg-details}. 

\section{Experiments}
We conduct three types of experiments to demonstrate the empirical superiority of proposed algorithms. As a starting point, we first show \plum{}'s computational efficiency by applying them to general prompt learning tasks with white-box models, where \plum{} produces nontrivial improvements over baselines within a fixed time limit. We further investigate \plum{} on prompt learning tasks with black-box models, where only discrete-input APIs are available. On these tasks, \plum{} also obtains significant performance gains with much fewer API calls. Eventually, we show that \plum{} is capable of discovering effective prompt patterns in both Chain-of-Thought and text-to-image generation tasks, making it a promising paradigm for boosting research in prompt learning.

\subsection{Plum for White-Box Prompt Learning}\label{evo-gen-prompt-white}

\paragraph{Dataset}
Our experiments are conducted on a subset of the Natural-Instructions datasets v2.6 release~\citep{mishra2022cross}, specifically focusing on eight binary classification tasks (task 019, task 021, task 022, task 050, task 069, task 137, task 139, and task 195). This dataset evaluates the instruction following ability of models, where proper prompt design plays an important role.

\paragraph{Baselines}
We compare our method with three popular prompt learning methods:
\begin{itemize}
    \item \textbf{BDPL}~\citep{diao2022black}: using variance-reduced policy gradient algorithm to estimate the gradients of parameters in the categorical distribution of each discrete prompt.
    \item \textbf{BBT}~\citep{sun2022black}: optimizing continuous prompts by covariance matrix adaptation evolution strategy.
    \item \textbf{GrIPS}~\citep{prasad2022grips}: generating discrete prompts by performing phrase-level operations, and selecting the best ones.
    \item \textbf{APO}~\citep{pryzant2023automatic}: automatically improve the initial prompt using beam search guided by textual ``gradient''.
\end{itemize}

\paragraph{Experiment Setup}
To make the comparison fair for all algorithms, we follow the same editing operators as GrIPS~\citep{prasad2022grips}, which generally generates discrete prompts by performing phrase-level operations and selecting the best ones.
We utilize GPT2-large~\citep{radford2019gpt2} for the backbone model of prompt learning. To accommodate computational resource limitations, we set the batch size to $1$ and the time-out threshold of the searching process to $45$ minutes for all methods. All methods are given a task-agnostic instruction: \textit{You will be given a task. Read and understand the task carefully, and appropriately answer [list\_of\_labels]}.
Here \textit{[list\_of\_labels]} serves as a placeholder to be replaced by an actual list of labels.

\paragraph{Results}
As demonstrated in Table \ref{tab:task-agnostic-instruction-only-white-box}, with limited time resources, \plum{} can provide generally on-par or better performance than baseline prompt learning methods. It is worth noticing that this is achieved even without the need for model gradients or any assistance from external LLMs, which shows the potential of \plum{} for offline prompt optimization scenarios where efficiency is of great importance. More experimental results are available in Appendix~\ref{appendix:extra-exp-white-box}.

\begin{table}[t]
  \begin{center}
    \begin{tabular}{lcr}
      \toprule

        Methods
        & Accuracy(\%)
        \\
        \midrule
        
        BBT~\citep{sun2022black}
        & 53.67$\pm$1.71$^\ddagger$
        \\
        
        BDPL~\citep{diao2022black}
        & 53.13$\pm$0.61
        \\
      
        GrIPS~\citep{prasad2022grips}
        & 53.67$\pm$0.87
        \\

        APO~\citep{pryzant2023automatic}
        &54.63$\pm$0.37
        \\
        
        \plumhc{}
        & 53.83$\pm$0.46
        \\
        
        \plumsa{}
        & 53.92$\pm$0.41
        \\
        
        \plumgam{}
        & 53.21$\pm$0.60
        \\
        
        \plumgac{}
        & 54.38$\pm$0.47
        \\  
        
        \plumts{}
        & 54.18$\pm$0.05
        \\    
        
        \plumhs{}
        & \textbf{55.04}$\pm$\textbf{0.56} 
        \\
        
        \bottomrule
    \end{tabular}
  \end{center}
  \caption{Impact of different search strategies with task-agnostic instruction and Instruction-Only prompts with gpt2-large as backbones. Here we set the time limit to 45 minutes for all methods.\\
  $^\ddagger$\quad We incorporate error bars for all reported results in this paper to illustrate the variability, where $\pm$ represents a plus or minus the standard deviation.}
  \label{tab:task-agnostic-instruction-only-white-box}
\end{table}

\subsection{Plum for Black-Box Prompt Learning}\label{evo-gen-prompt}

\paragraph{Experimental Setup} The experimental setup follows the same setting as white-box prompt learning, except for the backbone model, where we adopt GPT3-babbage instead. Notice that BBT~\citep{sun2022black} is no longer applicable in this setting since it requires continuous gradients in its prompt learning processes.

\begin{table*}[t]
  \begin{center}
    \begin{tabular}{cllllllr}
      \toprule

      Backbone
      & Methods
      & \makecell[c]{Max \\Iteration}
      & \makecell[c]{Batch \\Size}
      & Accuracy(\%)
      & \makecell[c]{API \\Calls}
      \\
      \midrule

      \multirow{6}{*}{\makecell[c]{GPT3-babbage\\ (without API Calls limit)}}

      & BDPL~\citep{diao2022black}
      & 50
      & 32
      & 57.38
      & 10000
      \\

      & GrIPS~\citep{prasad2022grips}
      & 10
      & 4
      & 54.41$\pm$0.55
      & 7740
      \\

      & \plumhc{}
      & 50
      & 20
      & 56.25$\pm$0.27
      & 10588
      \\

      & \plumsa{}
      & 50
      & 20
      & 54.92$\pm$0.78
      & 13900
      \\
      
      & \plumgam{}
      & 50
      & 20
      & 56.04$\pm$0.97
      & 10098
      \\
      
      & \plumgac{}
      & 50
      & 20
      & 56.63$\pm$0.97
      & 13367
      \\      
      
      & \plumts{}
      & 50
      & 20
      & 54.63$\pm$0.83
      & 1752
      \\   

      & \plumhs{}
      & 50
      & 20
      & \textbf{59.63}$\pm$\textbf{0.80}
      & 5494
      \\

      \midrule
      
      \multirow{5}{*}{\makecell[c]{GPT3-babbage\\ (with API Calls limit 8000)}}

      & BDPL
      & 40
      & 32
      & 56.25
      & --
      \\

      & GrIPS
      & 10
      & 4
      & 54.41$\pm$0.55
      & --
      \\

      & \plumhc{}
      & 50
      & 20
      & 55.21$\pm$0.48
      & --
      \\
        
      & \plumsa{}
      & 50
      & 20
      & 56.87$\pm$1.62
      & --
      \\
      
      & \plumgam{}
      & 50
      & 20
      & 55.63$\pm$0.47
      & --
      \\
      
      & \plumgac{}
      & 50
      & 20
      & 57.75$\pm$0.74
      & --
      \\  

      & \plumts{}
      & 50
      & 20
      & 54.92$\pm$1.26
      & --
      \\    

      & \plumhs{}
      & 50
      & 20
      & \textbf{59.75}$\pm$\textbf{0.81}
      & --
      \\ 

      \bottomrule
    \end{tabular}
  \end{center}
  \caption{Impact of different search strategies with task-agnostic instruction and Instruction-Only prompts with GPT3 as backbone.}
  \label{tab:task-agnostic-instruction-only}
\end{table*}

\paragraph{Results}
Table \ref{tab:task-agnostic-instruction-only} lists the performance of \plum{}-series algorithms, where under both scenarios of a limited number of iterations and API calls, \plumhs{} achieves the best performance. Specifically, \plumhs{} obtains this superior performance at a cost of much fewer API calls, rendering it a promising candidate for other prompt-tuning tasks. It is also worth noticing that given approximately the same amount of API calls, all \plum{} algorithms surpass GrIPS by a non-trivial margin.

\subsection{Plum for Prompt Discovery}\label{exp-plum-discover}

\paragraph{Text-to-Image Generation}
As demonstrated by images in Table~\ref{tab:plum_stable_diffusion}, \plum{} can also be applied to text-to-image generation tasks and obtain non-trivial improvements even without the assistance of external LLMs like GPT-3~\citep{brown2020gpt3}. By utilizing an offline image evaluator called PickScore~\citep{kirstain2023pick}, optimizing image relevance to the original prompt can be formulated as a black-box prompt learning problem, where a prompt's score is defined as the averaged PickScore of the images generated from the prompt. PickScore is a CLIP-based scoring function trained on a corpus of high-quality image text pair dataset. This makes improving the image quality quantitatively possible. For example, in the fourth image of Table~\ref{tab:plum_stable_diffusion}, where the image topic is ``cat in a library'', \plumhs{} augments the original prompt ``\textit{A cat prowling in a library at night, books and shadows, silent observer}'', and replaces it with a high-score prompt ``\textit{A Is that true? Not at all cat prowling in shadows And books shadows}''. Notice the resultant prompt highlights the key information ``cat, books, shadows'' and appends a prefix ``Is that true?'' with a sense of mystery. To the best of our knowledge, this pattern was unexplored before and can be beneficial for future prompt engineering development in this field. More experimental results and details are available in Appendix~\ref{appendix:prompt-cases}.

\newcommand{\scalesd}{0.8}
\begin{table*}[h!]
\centering
\begin{tabular}{cc}
\includegraphics[width=\scalesd\linewidth]{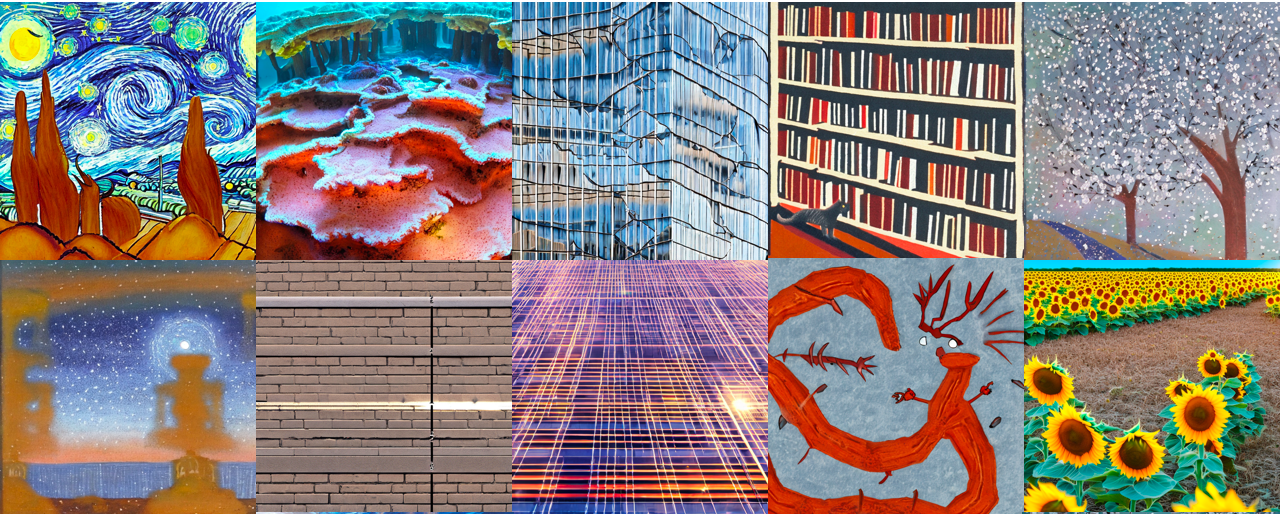}
\\
(a) SD-2.1 ($768\times768$ resolution, Average PickScore = 50.0)
\\
\includegraphics[width=\scalesd\linewidth]{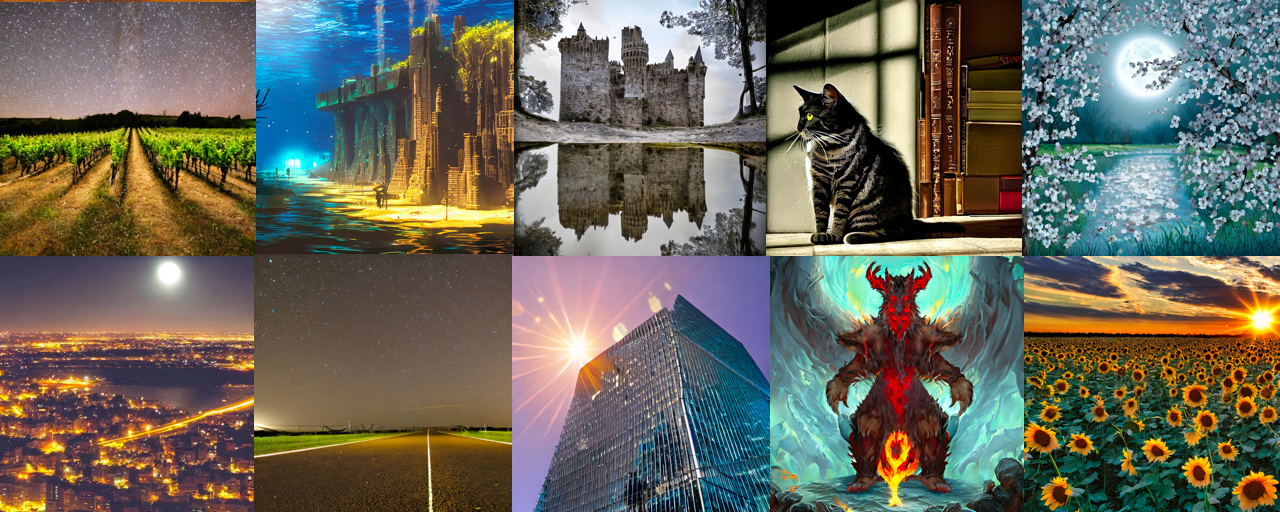}
\\
(b) \plumhs{} augmented SD-2.1 ($768\times768$ resolution, Average PickScore = 80.6)
\end{tabular}
\caption{\plumhs{} augmented prompts for \texttt{stable-diffusion-2-1}. Image topics: \textbf{Top}: 1) Starry night, 2) An underwater city, 3) Reflection, 4) Cat in a library, 5) Moonlight bathes blossom; \textbf{Bottom}: 1) Moonlight and city lights, 2) Boundary, 3) Sun glare on a skyscraper, 4) Fiery beast, 5) Sunflower.}
\label{tab:plum_stable_diffusion}
\end{table*}

\paragraph{Reasoning}
Another important feature of \plum{} is its compatibility with external LLM search operators for paraphrasing, which allows the discovery of interpretable prompt patterns. We apply this version of \plum{} to improve task-oriented LLMs' performance, where the detailed cases are presented in Table~\ref{tab:case-examples} of Appendix~\ref{appendix:prompt-cases}. Via studying the characteristics of the searched prompts, we found that the performance of initial prompts can be boosted by further completion of the logical chain. This way, the logical coherence of prompts can be improved while still retaining their intended meaning. For instance, \textit{`If there are 3 cars in the parking lot and 2 more cars arrive, how many cars are in the parking lot (total number of cars in the parking lot + 2)'}, where GPT-3 completes the content inside the parentheses, leading to an easier Chain-of-Thought reasoning for LLMs.

Furthermore, it is observed that the quality of the prompts can be enhanced by incorporating a more comprehensive and lucid explanation of the statement. To illustrate, in a case of prompting for AQuA, GPT3.5 provides an initial explanation of the term \textit{`keystrokes'} with \textit{`(the act of pressing a key on a keyboard)'} to ensure clarity, and further gives a precise description for the range \textit{`from 1 to 500'} with \textit{`(inclusive)'}. This suggests that knowledge-augmented prompt learning methods can be beneficial for black-box prompt tuning. 

\section{Discussion}

\paragraph{\plum{} with LLM-driven Mutators}

LLM-based prompt tuning methods are emerging as one of the most competitive approaches in prompt learning. Recent works like \citet{pryzant2023automatic, guo2023evoprompt} have shown significant performance improvements through utilizing the LLM-driven prompt mutators, highlighting the success of such approaches. Compared with them, \plum{} offers a complementary and orthogonal paradigm. For example, by incorporating the LLM-based operators in \plum{} algorithms, our Chain-of-Thought task experiments in Appendix~\ref{appendix:cot-exps} demonstrate effectiveness in generating high-quality, interpretable few-shot-CoT prompts for reasoning tasks (as shown in Table~\ref{tab:case-examples}). This suggests that LLMs can be a powerful tool for prompt exploration within the \plum{} paradigm, particularly when computational resources are abundant.

\paragraph{Strengths of Different \plum{} Algorithms}

In both white-box and black-box settings, our experiments demonstrate that \plumhs{} achieves significant performance gains on prompt tuning tasks against baselines and the other \plum{} algorithms. This is because it takes into account the composability of different prompt segments, which can be seen as an extension of evolutionary algorithms. Notably, in tasks like image generation (detailed in Section~\ref{exp-plum-discover}), we observe that updates to different segments can occur in parallel, highlighting this approach's efficiency.

On the other hand, while two baseline methods may have 
superior performance over most other \plum{} algorithms, those metaheuristics possess nice properties that can be beneficial for the community. 

\begin{itemize}

    \item \plumhc{} stands out for its simplicity and efficiency among \plum{} family. It shares the same form of GrIPS~\citep{prasad2022grips} and is one of the most commonly adopted paradigms in LLM-based methods. \plumhc{} can serve as a foundational algorithm for developing new mutators. Its straightforward nature allows for easier debugging and comparison when developing more complex algorithms in the \plum{} family.

    \item For \plumsa{}, it boasts a quite simple form, which only differs one line from \plumhc{}. Crucially, it guarantees to find the global optimum with the infinite time horizon. This characteristic makes it valuable for future theoretical analyses of discrete prompt learning.
    
    \item \plumgam{} and \plumgac{}, the genetic algorithm, can be considered one of the most widely adopted algorithms in evolutionary search, whose results can be useful for future comparisons on the same tasks for other researchers. Additionally, our experiments also demonstrate that these algorithms achieve superior performance compared to baseline approaches.
    
    \item For \plumts{}, although its performance is not the best, it achieves on-par performance as GrIPS in ~4 times fewer API calls (as shown in Table \ref{tab:task-agnostic-instruction-only}). This advantage makes \plumts{} an extremely favorable choice for applications with limited budgets or API call quotas.
    
\end{itemize}

\section{Conclusion}

In this paper, a novel paradigm of prompt learning is proposed. By formulating prompt learning as a black-box discrete optimization problem, we are able to apply various metaheuristics algorithms to help discover effective prompts, while guaranteeing the whole process to be automatic, discrete, gradient-free, interpretable, and applicable to black-box models. Six typical \plum{} algorithms, Hill Climbing, Simulated Annealing, Genetic Algorithm (Mutation Only), Genetic Algorithm (with Crossover), Tabu Search, and Harmony Search all obtain non-trivial improvements over GrIPS on instruction following tasks. Furthermore, with the assistance of metaheuristics-based prompt learning, we are capable of discovering new patterns of effective prompts unexplored in the past.

\section*{Limitations}
As an early step of bridging metaheuristics and prompt learning in a unified paradigm, we only implement six algorithms to prove the validity of this concept. Nevertheless, there are still a bunch of other potential metaheuristics that can be incorporated into this framework, producing more empirically useful methods for solving real-world problems. We hope \plum{} is just a starting point, more metaheuristics, such as Particle Swarm Optimization~\citep{kennedy1995particle}, Ant Colony Optimization~\citep{dorigo1997ant} and Cuckoo search~\citep{yang2009cuckoo} can be combined with more neighborhood definitions, like retrieval-based completions and multimodal, to inspire more practical algorithms, which can eventually help us find a way to properly communicate with Large Language Models.

The potential risks of this work are the general risks of black prompt tuning methods, where more powerful prompt learning algorithms enable more powerful prompt attack approaches. Nevertheless, the silver lining is that most prompt attack methods normally in term generate adversarial samples that help improve the robustness of models, which can be considered beneficial in the long term.

\bibliography{main}
\bibliographystyle{plainnat}

\appendix

\newcommand{\green}[1]{\colorbox{ForestGreen}{\textcolor{white}{\texttt{#1}}}}
\newcommand{\red}[1]{\colorbox{red}{\textcolor{white}{\texttt{#1}}}}
\newcommand{\orange}[1]{\colorbox{orange}{\textcolor{white}{\texttt{#1}}}}
\newcommand{\tblue}[1]{\textcolor{blue}{#1}}
\newcommand{\tblack}[1]{#1}

\newpage

\section{Discovered Prompt Patterns}
\label{appendix:prompt-cases}

\paragraph{Text-to-Image Generation}

Several detailed prompts in text-to-image generation tasks are presented in Table~\ref{tab:extra_plum_stable_diffusion}. It can be observed that those improved prompts have a tendency to focus more on key descriptive elements in the image, such as ``shadows'', ``shimmer in petals'' or ``sunset''. The augmented prompts also use ``There is'' or ``They have'' to emphasize the main theme of the image and remove other distractive words.

\newcommand{\scalesdappendix}{1.0}
\begin{table*}[h!]
\centering
\begin{tabular}{p{2cm}p{3cm}p{3cm}p{3cm}p{3cm}}
\toprule
Topic & Original Image & Resultant Image & Original Prompt & Augmented Prompt
\\
\midrule
\vspace*{-3cm} Cat in a library
& \includegraphics[width=\scalesdappendix \linewidth]{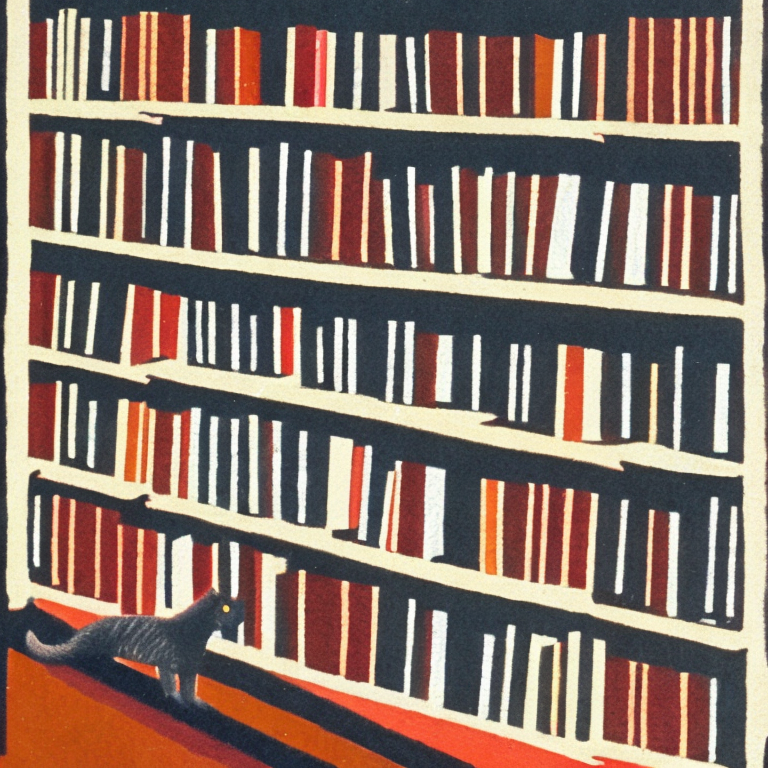}
& \includegraphics[width=\scalesdappendix \linewidth]{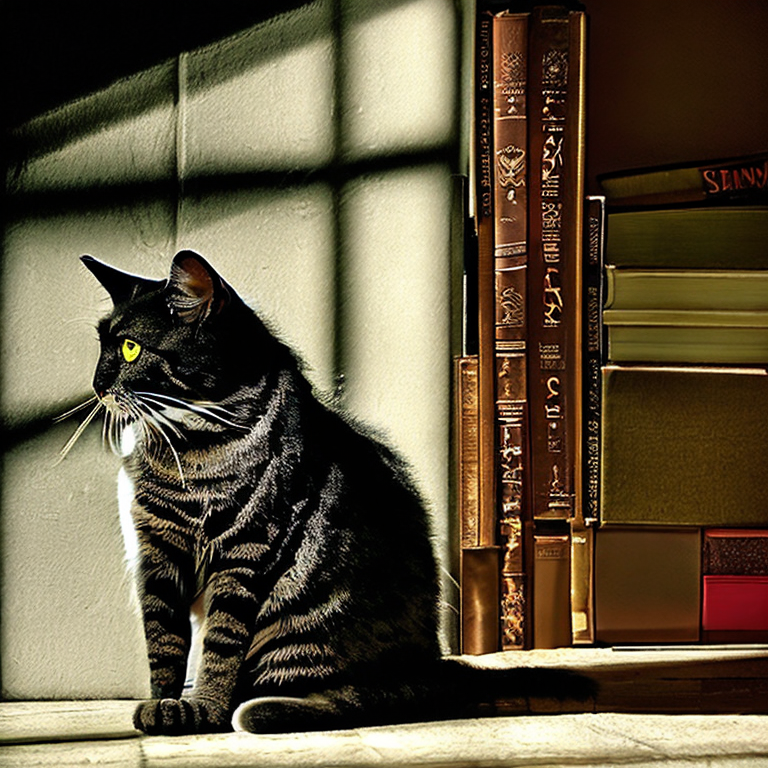}
& \vspace*{-3cm}A cat prowling in a library at night, books and shadows, silent observer.
& \vspace*{-3cm}A Is that true? Not at all cat prowling in shadows And books shadows,
\\
\midrule
\vspace*{-3cm} Moonlight bathes blossom
& \includegraphics[width=\scalesdappendix \linewidth]{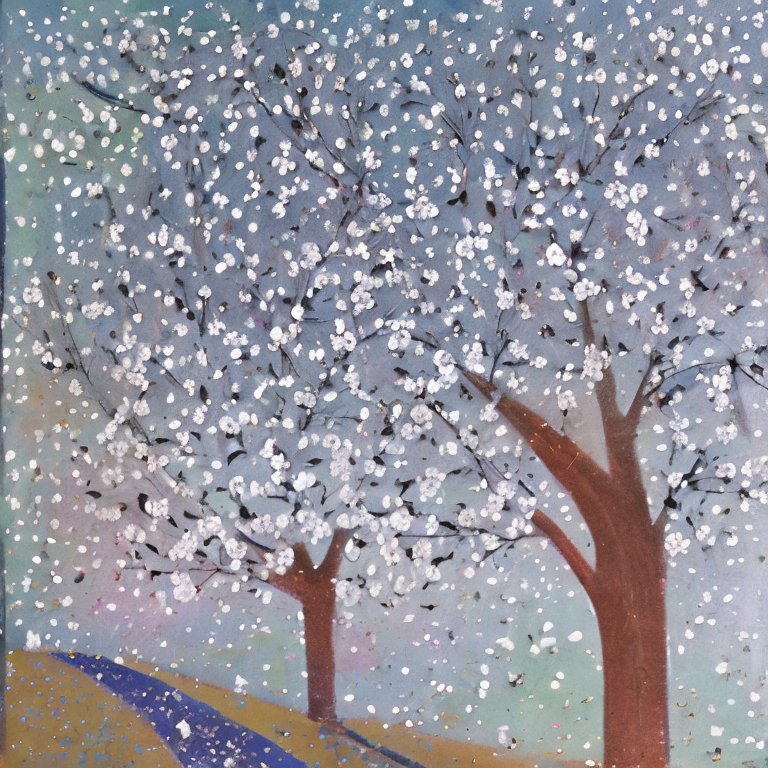}
& \includegraphics[width=\scalesdappendix \linewidth]{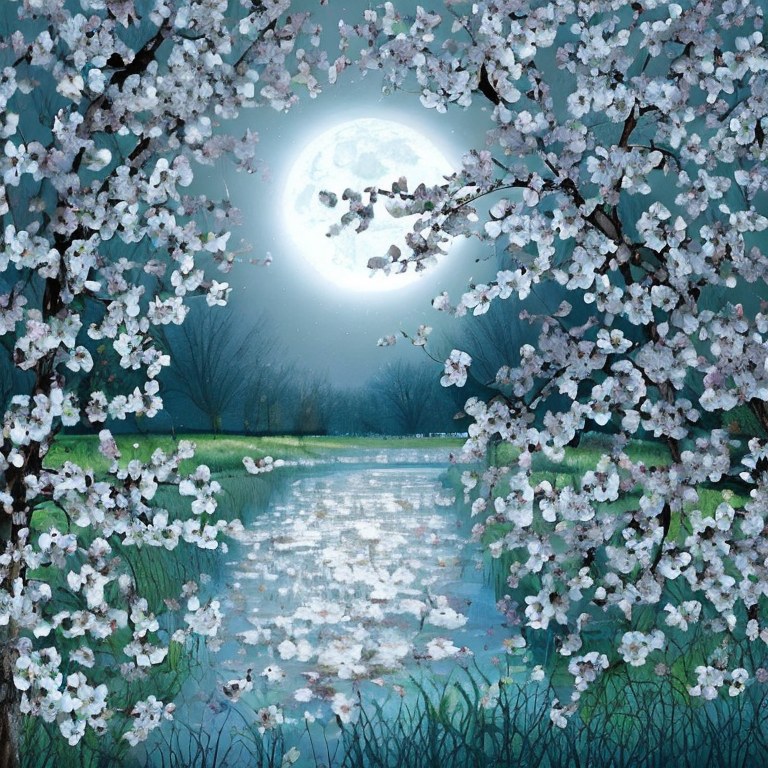}
& \vspace*{-3cm} Moonlight bathes blossom grove, petals shimmer in silvery luminescence
& \vspace*{-3cm} Moonlight bathes blossom silvery There is shimmer in petals
\\
\midrule
\vspace*{-3cm} Sunflower
& \includegraphics[width=\scalesdappendix \linewidth]{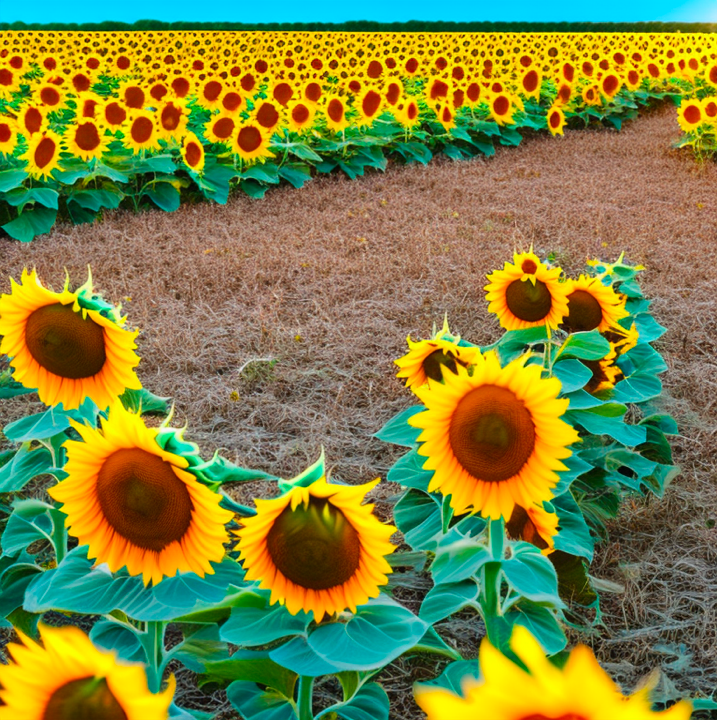}
& \includegraphics[width=\scalesdappendix \linewidth]{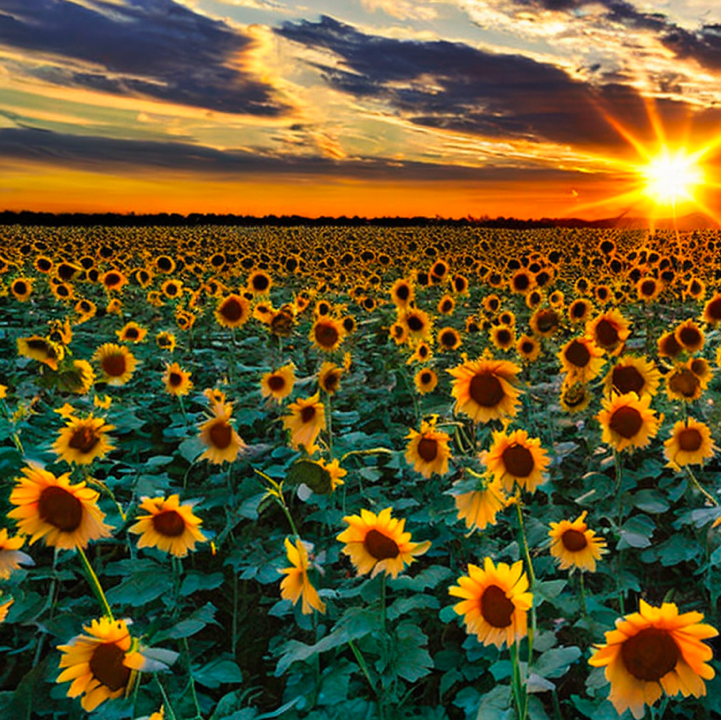}
& \vspace*{-3cm} Sunflower field at sunset, golden hour, endless summer.
& \vspace*{-3cm} Sunflower endless field sunset, at sunset,
\\
\midrule
\vspace*{-3cm} Moonlight bathes blossom
& \includegraphics[width=\scalesdappendix \linewidth]{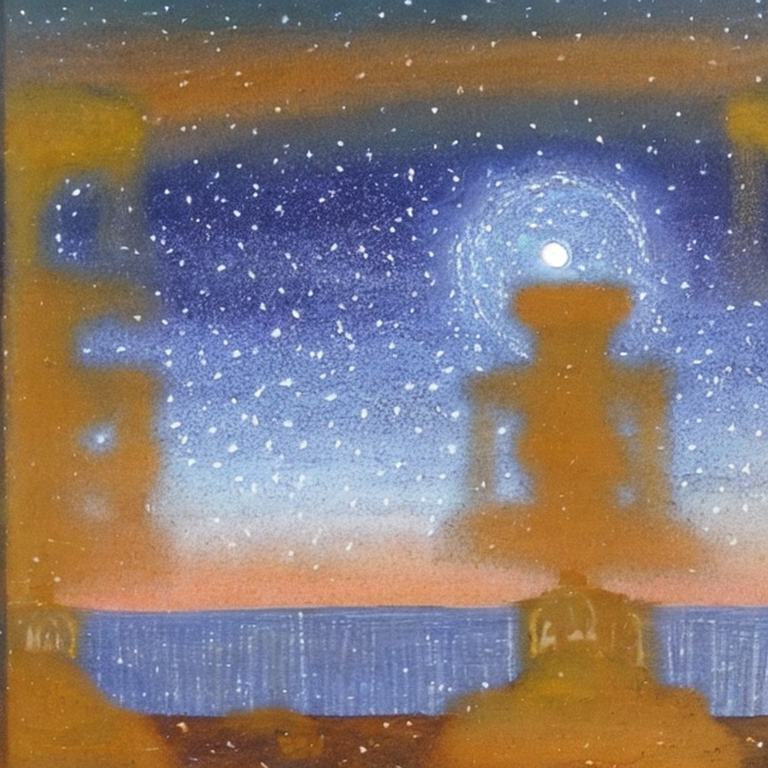}
& \includegraphics[width=\scalesdappendix \linewidth]{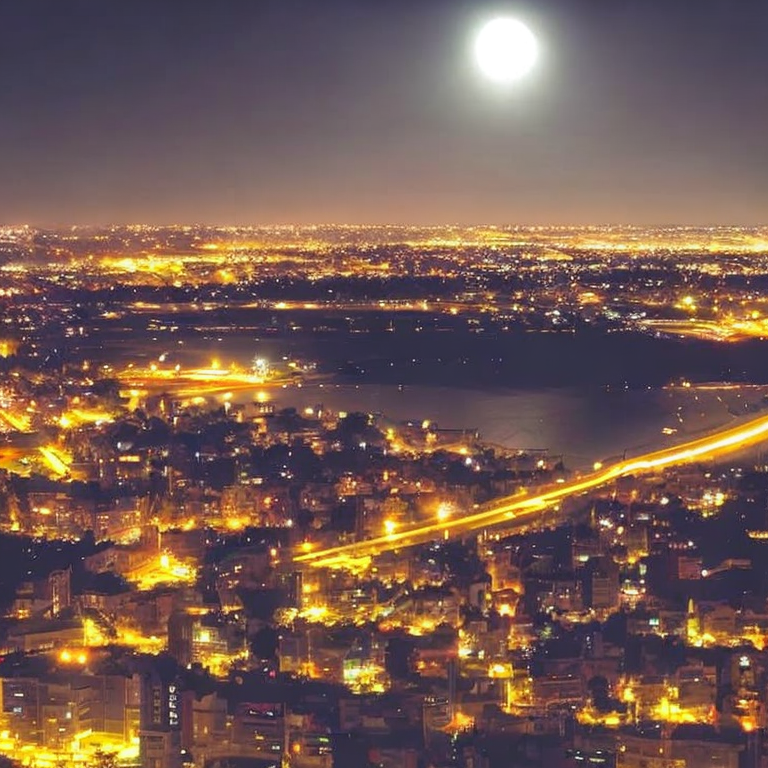}
& \vspace*{-3cm} Moonlight illuminating at night, distant city lights.
& \vspace*{-3cm} Moonlight illuminating at night distant city They have lights
\\
\midrule
\vspace*{-3cm} Glowing Mushroom Forest
& \includegraphics[width=\scalesdappendix \linewidth]{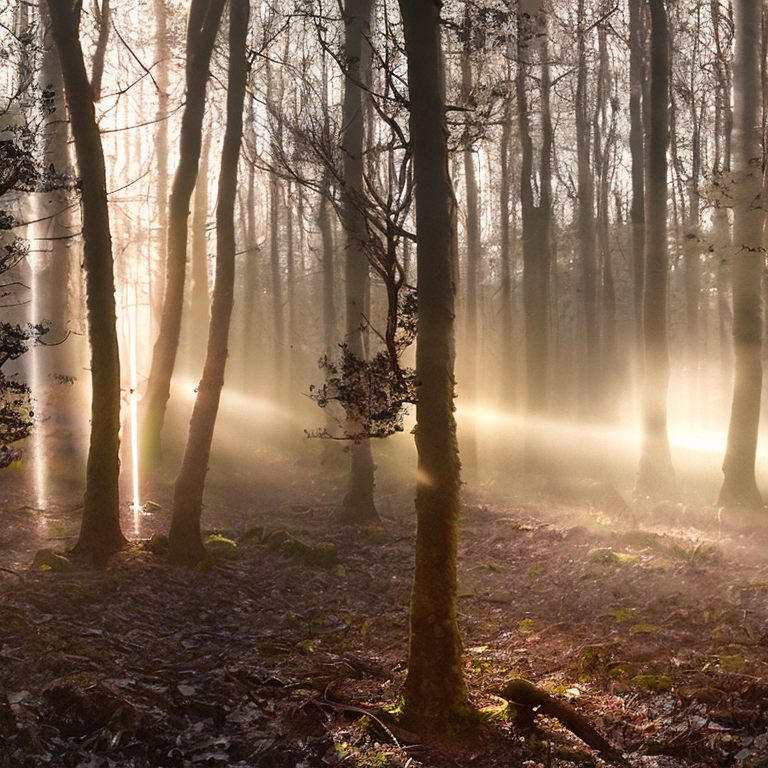}
& \includegraphics[width=\scalesdappendix \linewidth]{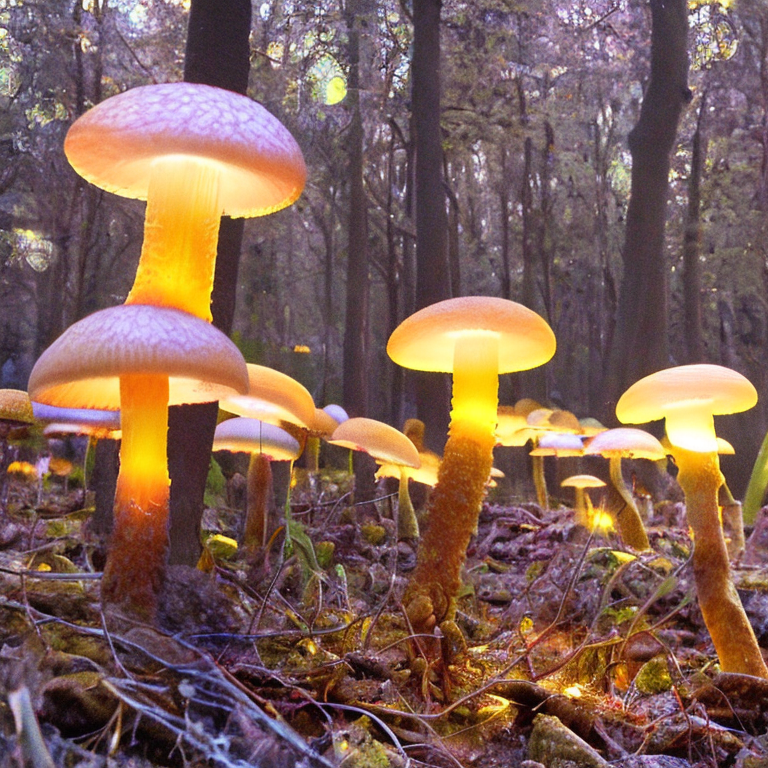}
& \vspace*{-3cm} Illuminated fungi cast ethereal light in dense woodland
& \vspace*{-3cm} Illuminated fungi cast It's something The and It is ethereal in dense
\\
\midrule
\end{tabular}
\caption{\plumhs{} augmented prompts for \texttt{stable-diffusion-2-1}}.
\label{tab:extra_plum_stable_diffusion}
\end{table*}

\paragraph{Reasoning}
Here we present several typical cases of discovered prompt patterns for reasoning tasks, as shown in Table~\ref{tab:case-examples}. It turns out that 1) completing the logical chain, 2) explaining the meaning of a term, or 3) providing additional clarification of a statement all help LLMs improve performance in reasoning tasks.

\begin{table*}[t!]
\footnotesize
    \begin{tabular}{m{2cm} m{13cm} } 
        \toprule
        Tasks & Searched Prompts\\
        \midrule
        \multirow{2}{*}{GSM8K} 
        & 
        \textbf{Initial:} $\cdots$ If there are 3 cars in the parking lot and 2 more cars arrive, how many cars are in the parking lot? $\cdots$ \\
        &
        \textbf{GPT3:}  $\cdots$ If there are 3 cars in the parking lot and 2 more cars arrive, how many cars are in the parking lot \colorbox{yellow!20}{(total number of cars in the parking lot + 2)}? $\cdots$\\

        \midrule
        \multirow{2}{*}{ASDiv} 
        &
        \textbf{Initial:} $\cdots$ Leah had 32 chocolates and her sister had 42. If they ate 35, how many pieces do they have left in total?  $\cdots$ \\
        &
        \textbf{GPT3.5:} $\cdots$ Leah had 32 chocolates and her sister had 42  \colorbox{yellow!20}{(total number of chocolates owned by Leah} \colorbox{yellow!20}{and her sister)}. If they ate 35 \colorbox{yellow!20}{(total number of chocolates eaten by Leah and her sister)}, how many pieces do they have left in total \colorbox{yellow!20}{(total} \colorbox{yellow!20}{number of chocolates remaining with Leah and her sister)}?  $\cdots$ \\

         \midrule
        \multirow{2}{*}{AQuA} 
        &
        \textbf{Initial:} $\cdots$ How many keystrokes are needed to type the numbers from 1 to 500? Answer Choices: (a) 1156 (b) 1392 (c) 1480 (d) 1562 (e) 1788.$\cdots$\\
        &
        \textbf{GPT3.5:} $\cdots$ How many keystrokes \colorbox{yellow!20}{(the act of pressing a key on a keyboard)} are needed to type the numbers from 1 to 500 \colorbox{yellow!20}{(inclusive)}? Answer Choices: (a) 1156 (b) 1392 (c) 1480 (d) 1562 (e) 1788.$\cdots$\\
        
        \midrule
        \multirow{3}{*}{CSQA} 
        &
        \textbf{Initial:} $\cdots$ What do people use to absorb extra ink from a fountain pen? Answer Choices: (a) shirt pocket (b) calligrapher’s hand (c) inkwell (d) desk drawer (e) blotter $\cdots$\\
        &
        \textbf{GPT3.5:} $\cdots$ What do people use to absorb extra ink \colorbox{yellow!20}{(liquid used for writing or printing)} from a fountain pen \colorbox{yellow!20}{(a pen that uses a reservoir of liquid ink to write)}? Answer Choices: (a) shirt pocket (b) calligrapher’s hand (c) inkwell (d) desk drawer (e) blotter (a piece of absorbent material used to soak up excess ink or to dry freshly written ink) $\cdots$ \\
        \midrule
        
        \multirow{4}{*}{StrategyQA}
        &
        \textbf{Initial:} $\cdots$ Yes or no: Hydrogen’s atomic number squared exceeds number of Spice Girls? $\cdots$\\
        &
        \textbf{GPT3.5:} $\cdots$ Yes or no: Does hydrogen have an atomic number of 1 \colorbox{yellow!20}{(the number of protons in the} \colorbox{yellow!20}{nucleus of an atom)}?$\cdots$\\
         \bottomrule 
\end{tabular}\caption[]{Examples of initially given prompts and the corresponding searched prompts searched by Evo-CoT}\label{tab:case-examples}
\end{table*}

\section{Additional Experimental Results}

\subsection{Plum for White-Box Prompt Learning}
\label{appendix:extra-exp-white-box}

\begin{figure}[t]
    \centering
    \begin{subfigure}{0.22\textwidth}
      \includegraphics[width=1\linewidth]{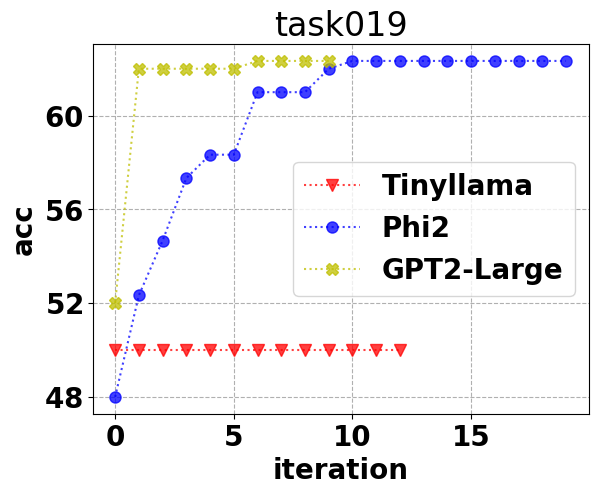}
      \caption{Task 019}
      \label{fig:iter-task019}
    \end{subfigure}%
    \begin{subfigure}{0.23\textwidth}
      \includegraphics[width=1\linewidth]{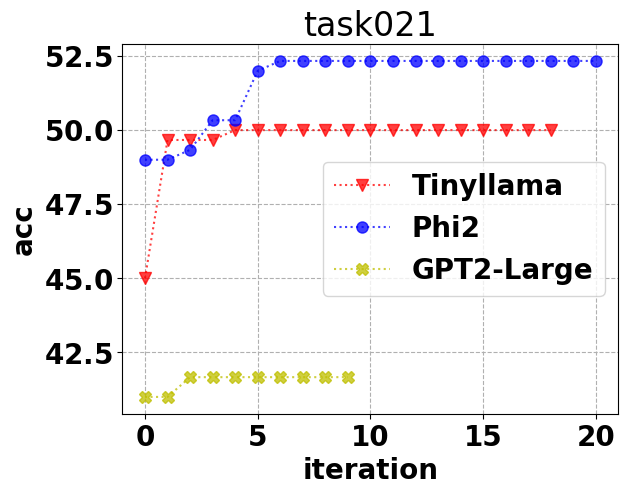}
      \caption{Task 021}
      \label{fig:iter-task021}
    \end{subfigure}%
    \begin{subfigure}{0.22\textwidth}
      \includegraphics[width=1\linewidth]{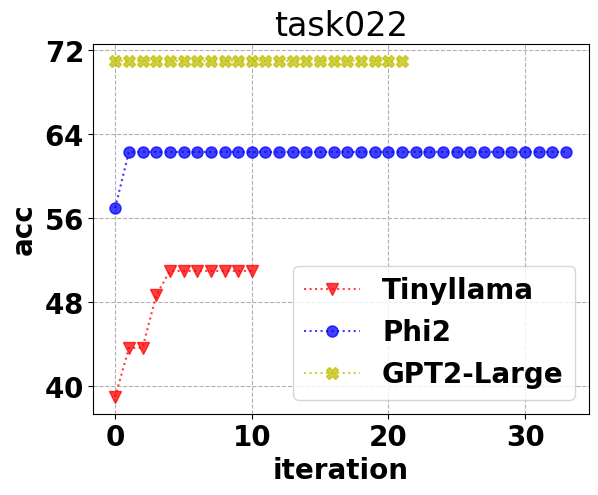}
      \caption{Task 022}
      \label{fig:iter-task022}
    \end{subfigure}%
    \begin{subfigure}{0.22\textwidth}
      \includegraphics[width=1\linewidth]{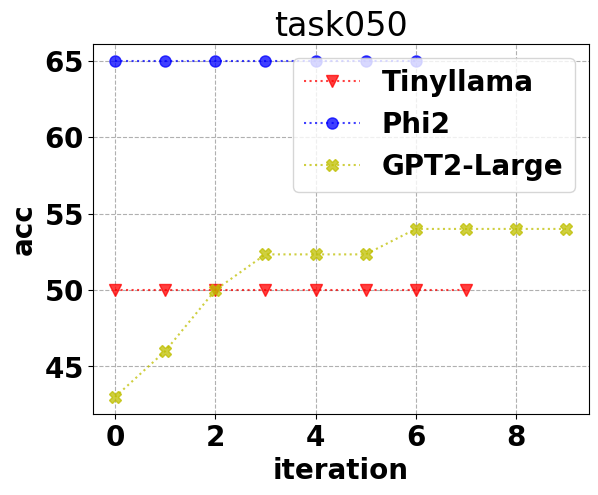}
      \caption{Task 050}
      \label{fig:iter-task050}
    \end{subfigure}%
    \\
    \begin{subfigure}{0.22\textwidth}
      \includegraphics[width=1\linewidth]{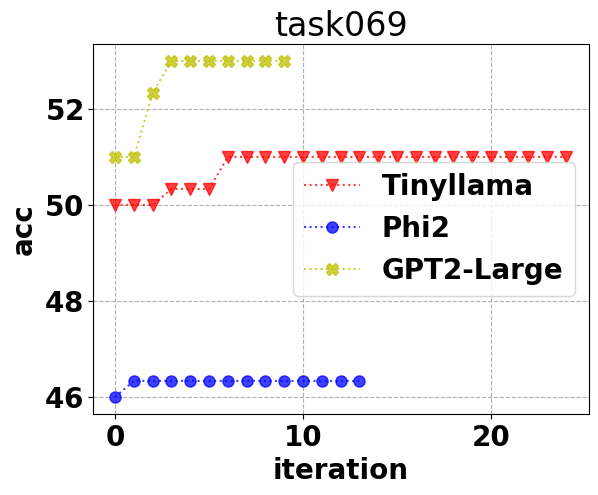}
      \caption{Task 069}
      \label{fig:iter-task069}
    \end{subfigure}%
    \begin{subfigure}{0.23\textwidth}
      \includegraphics[width=1\linewidth]{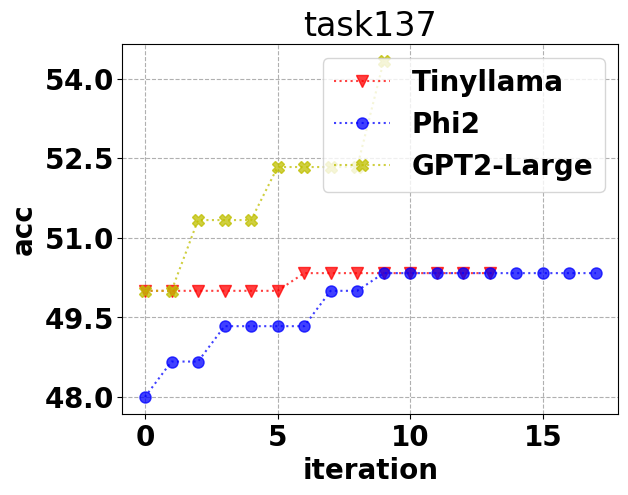}
      \caption{Task 137}
      \label{fig:iter-task137}
    \end{subfigure}%
    \begin{subfigure}{0.23\textwidth}
      \includegraphics[width=1\linewidth]{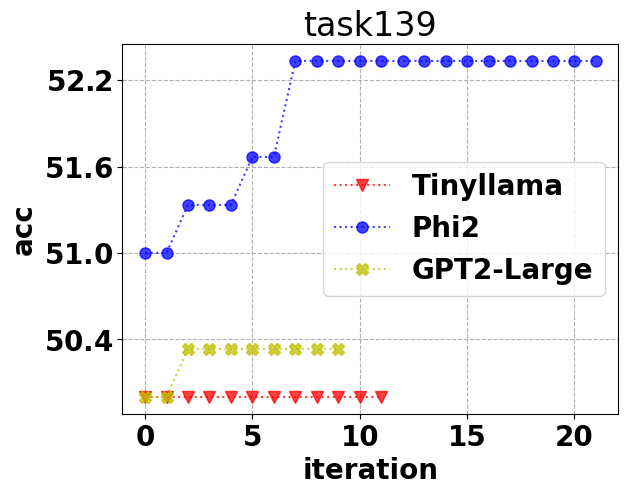}
      \caption{Task 139}
      \label{fig:iter-task139}
    \end{subfigure}%
    \begin{subfigure}{0.22\textwidth}
      \includegraphics[width=1\linewidth]{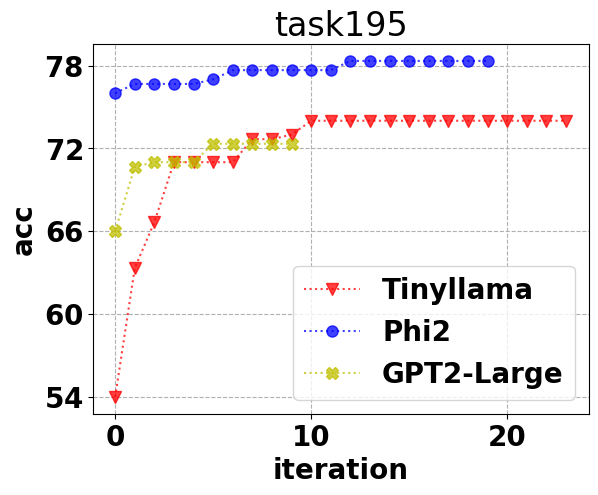}
      \caption{Task 195}
      \label{fig:iter-task195}
    \end{subfigure}%
    \caption{Performance of \plumhs{} across different iterations on subtasks of Natural-Instructions dataset.}
    \label{fig:iter-hs}
\end{figure}

More experimental results on Phi-2~\citep{microsoft2024phi2} and TinyLlama~\citep{zhang2024tinyllama} are shown in Table~\ref{tab:extra-white-box}, where we still observe on-par or better performance when compare \plum{} with baseline methods. Additionally, Figure \ref{fig:iter-hs} illustrates the performance (accuracy) of \plumhs{} on the Natural-Instructions dataset subtasks as the number of iterations increases, highlighting the effectiveness and efficiency of our proposed paradigm.

\begin{table}[t]
  \begin{center}
    \begin{tabular}{lcr}
      \toprule

        Methods
      & Phi-2 & TinyLlama
      \\
      \midrule

      BDPL ~\citep{diao2022black}
      & 53.88$\pm$1.02 &  51.08$\pm$0.58
      \\
      
      GrIPS ~\citep{prasad2022grips}
      &54.41$\pm$0.32 & 51.17$\pm$0.78
      \\

      APO~\citep{pryzant2023automatic}
      &55.04$\pm$0.12 & 47.58$\pm$0.05
      \\

      \plumhc{}
      & 54.46$\pm$1.55 & 51.71$\pm$0.77
      \\
        
      \plumsa{}
      & 54.54$\pm$1.50 & 52.25$\pm$1.24
      \\
      
      \plumgam{}
      & 53.67$\pm$0.79 & 51.13$\pm$1.24
      \\
      
      \plumgac{}
       & 54.38$\pm$1.08 & 52.08$\pm$0.33
      \\  

      \plumts{}
       & 54.25$\pm$1.89 & \textbf{52.38}$\pm$\textbf{1.64}
      \\    

      \plumhs{} & \textbf{55.25}$\pm$\textbf{0.91} & 51.54$\pm$0.35
      \\
      
      \bottomrule
    \end{tabular}
  \end{center}
  \caption{Additional Experiments on 
white-box prompt learning tasks. BBT~\citep{sun2022black}'s codebase is not fully compatible with Phi-2 and TinyLlama, hence the corresponding result is not available.}
  \label{tab:extra-white-box}
\end{table}

\subsection{Plum for Chain-of-Thought}\label{appendix:cot-exps}

The \plum{} paradigm offers a natural approach to obtaining high-quality few-shot-CoT prompts for improving LLM's reasoning performances.
To ensure that this class of prompts still remains logical and interpretable after optimization, we have made further improvements to the search process by introducing an additional LLM. Specifically, we encourage the language models to provide additional information and details in the \texttt{add} operation and introduce a teaching mechanism to guide the language models in revising prompts. On top of that, we empower the language models to autonomously split the prompts into phrases based on their own comprehension of the underlying text structure.

\begin{table*}[h!]\centering
    \scalebox{0.88}{
    \begin{tabular}{cllllllllllll}
    \toprule
    & Prompting 
    & GSM8K 
    & ASDiv 
    & SVAMP
    & AQuA 
    & CSQA 
    & StrategyQA \\
    \midrule
    \multirow{2}{*}{\makecell[c]{GPT3 \\ (\emph{text-curie-001})}}

    & CoT
    & 0.00
    & 7.17
    & 6.67
    & 20.83
    & 24.00
    & 52.33 \\

    & \plumgam{}*
    & 4.66$\pm$0.47
    & 9.67$\pm$1.25
    & 11.67$\pm$0.94
    & 22.00$\pm$0.82
    & 29.00$\pm$0.82
    & 61.00$\pm$1.63\\

    & \plumhs{}*
    & 4.33$\pm$0.47
    & 12.33$\pm$1.25
    & 11.00$\pm$0.82
    & 21.66$\pm$0.47
    & 30.00$\pm$0.82
    & 54.00$\pm$0.00\\
    
    \midrule
    \multirow{2}{*}{\makecell[c]{GPT3.5 \\ (\emph{gpt-3.5-turbo-0301)}}}

    & CoT 
    & 85.00
    & 92.00  
    & 85.17
    & 59.83 
    & 73.17
    & 72.50 \\
    
    & \plumgam{}*
    & 87.33$\pm$0.94
    & 94.67$\pm$0.94
    & 88.00$\pm$0.82
    & 67.67$\pm$1.25
    & 77.33$\pm$0.47
    & 78.00$\pm$1.41 \\

    & \plumhs{}*
    & 85.33$\pm$0.47
    & 92.33$\pm$0.47
    & 87.00$\pm$0.82
    & 67.67$\pm$1.89
    & 77.00$\pm$0.82
    & 73.00$\pm$0.82 \\
    
    \bottomrule 
    
    \end{tabular}}
    \caption{Improvements of optimization performance for the few-shot-CoT prompts searched by \plumgam{}* and \plumhs{}* as the searching strategies on math word solving and commonsense reasoning tasks. }
  \label{tab:few-shot-cot-improv}
\end{table*}

\begin{figure}[t]
    \centering
    \includegraphics[scale=0.45]{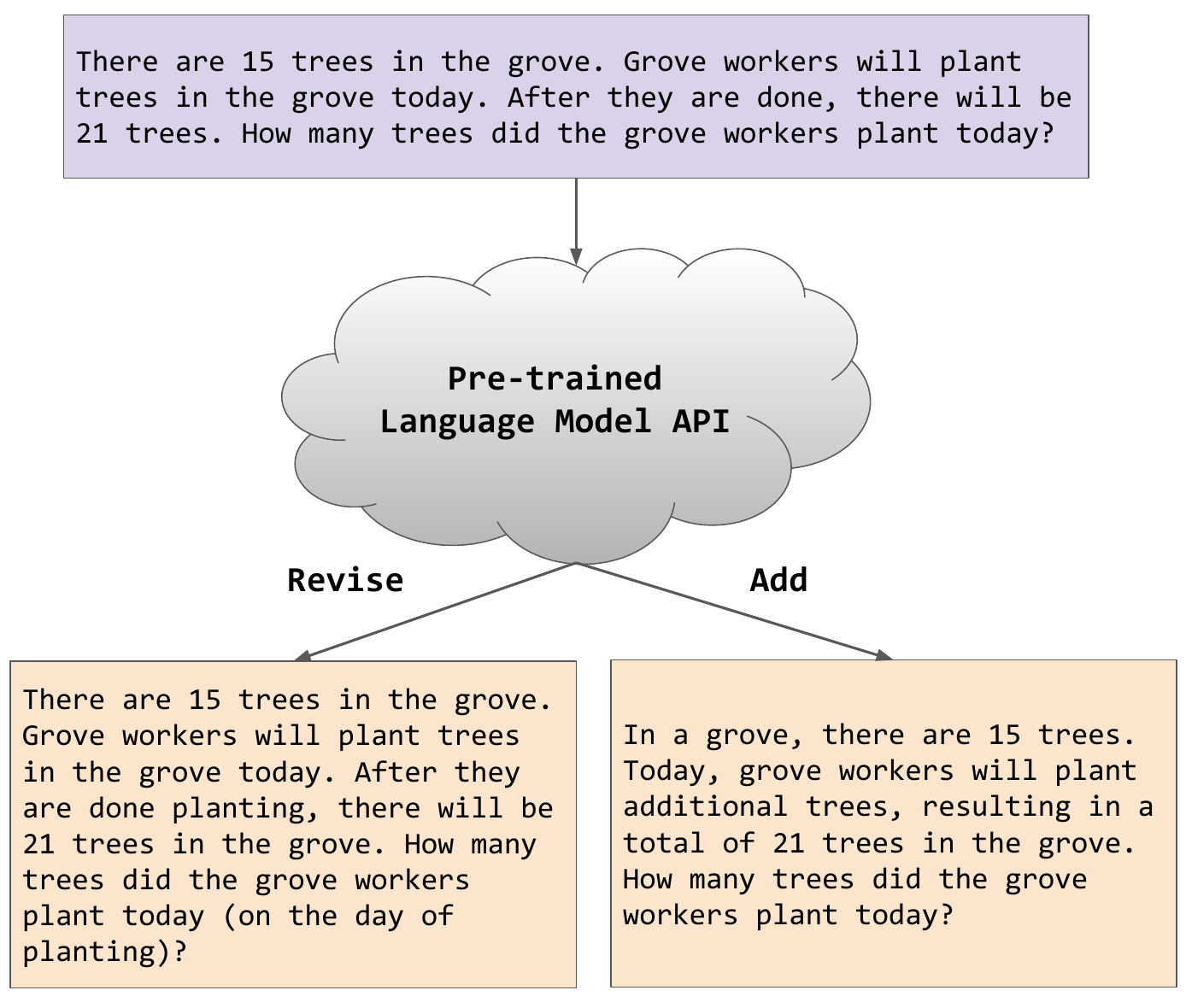}
    \caption{Illustration of the improved \texttt{add} operation and revision by language models}
    \label{fig:add-revise}
\end{figure}

As depicted in Figure~\ref{fig:add-revise}), these enhancements utilize the power of modern LLMs, aiming to accelerate the searching process of interpretable and target-model-friendly prompts via improved neighborhood definitions.
\paragraph{Dataset}
We test our methods on math word solving and commonsense reasoning tasks, including GSM8k~\citep{Cobbe2021Oct}, ASDiv~\citep{miao-etal-2020-diverse}, SVAMP~\citep{Patel2021Mar}, AQuA~\citep{koncel-kedziorski-etal-2016-mawps},
CSQA~\citep{talmor-etal-2019-commonsenseqa}, and StrategyQA~\citep{Geva2021Dec}. Those tasks evaluate models' reasoning ability, where prompts are expected to elicit their logical thinking skills. 

\paragraph{Experiment Setup}
\plum{} is applied to improving the performance of two backbones: GPT3 (text-curie-001) and GPT3.5 (gpt-turbo-3.5-0301) on aforementioned five reasoning tasks. Specifically, we employ \plumgam{}* and \plumhs{}* for this setting, majorly based on two reasons:
1) The effectiveness of both algorithms has been demonstrated in the previous general prompt learning setting, making them reasonable candidates for further extensions;
2) Compared with \plumgac{}, both algorithms have lower computational complexity and similar exploration ability. 
Moreover, considering the API budgets and the extensive size of the test set for both math word solving and commonsense reasoning tasks, we chose to exclusively showcase the optimization performance of \plumgam{}* and \plumhs{}* on the training sets for reference.

\paragraph{Results}
In this section, we present the enhancement achieved by the few-shot-CoT prompts searched by \plum{} across five demanding downstream tasks, as shown in Table \ref{tab:few-shot-cot-improv}. These results clearly demonstrate the superiority of \plumgam{}* and \plumhs{}* over standard few-shot-CoT prompts used for GPT3 and GPT3.5. In particular, our methods achieve universal improvements across those five tasks on almost all the backbone models.

In addition, despite the already impressive performance of GPT3.5 with standard few-shot-CoT prompts, we continue to observe even greater average improvements compared to GPT3. This is due to two key factors. Firstly, the edit operations we have developed in this section allow GPT3.5 to generate better prompts for itself. Secondly, GPT3.5 exhibits stronger capabilities in comprehending and leveraging the enhanced prompts, further improving its overall performance.

\begin{figure}[h]
    \centering
    \includegraphics[scale=0.25]{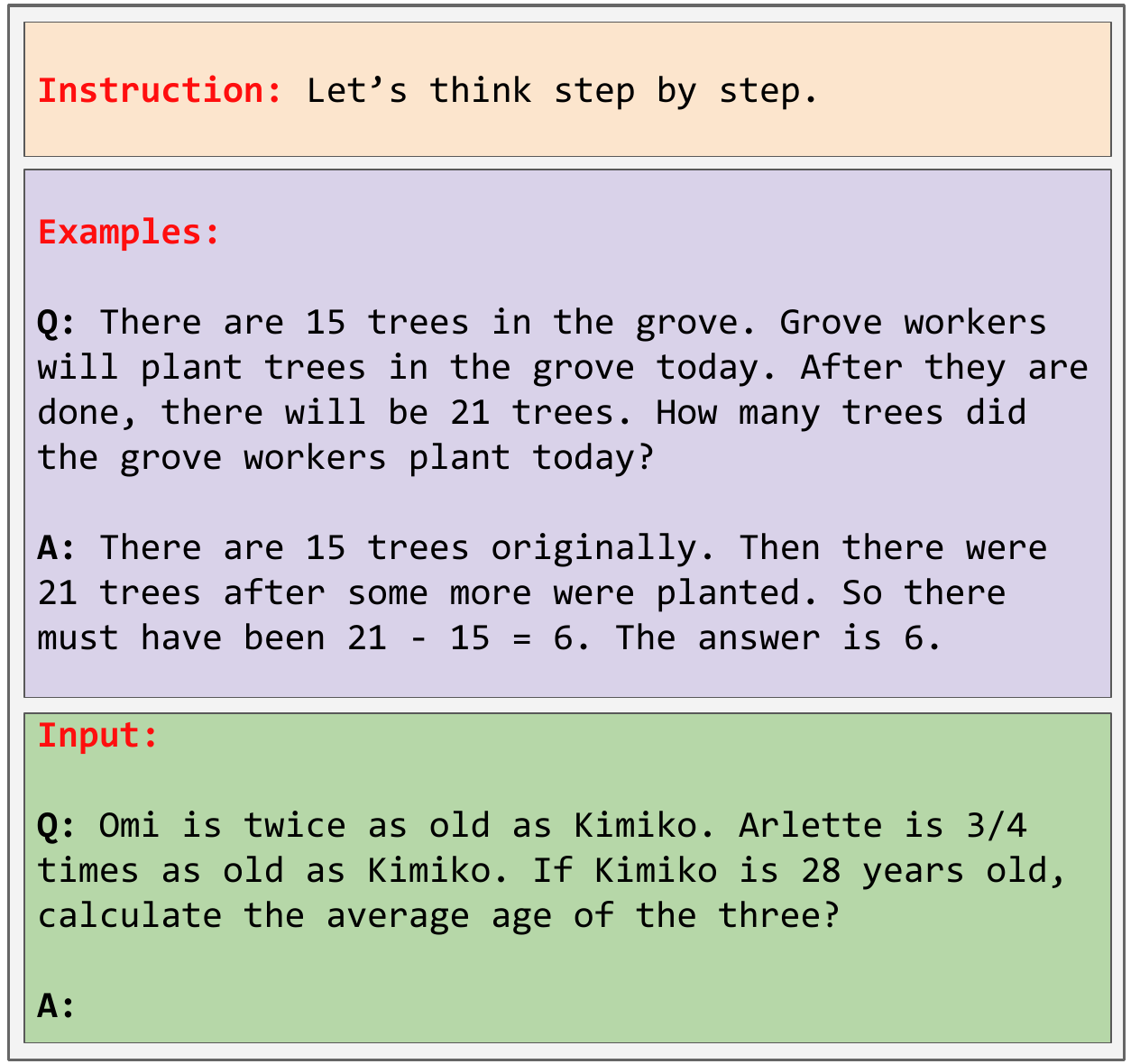}
    \includegraphics[scale=0.25]{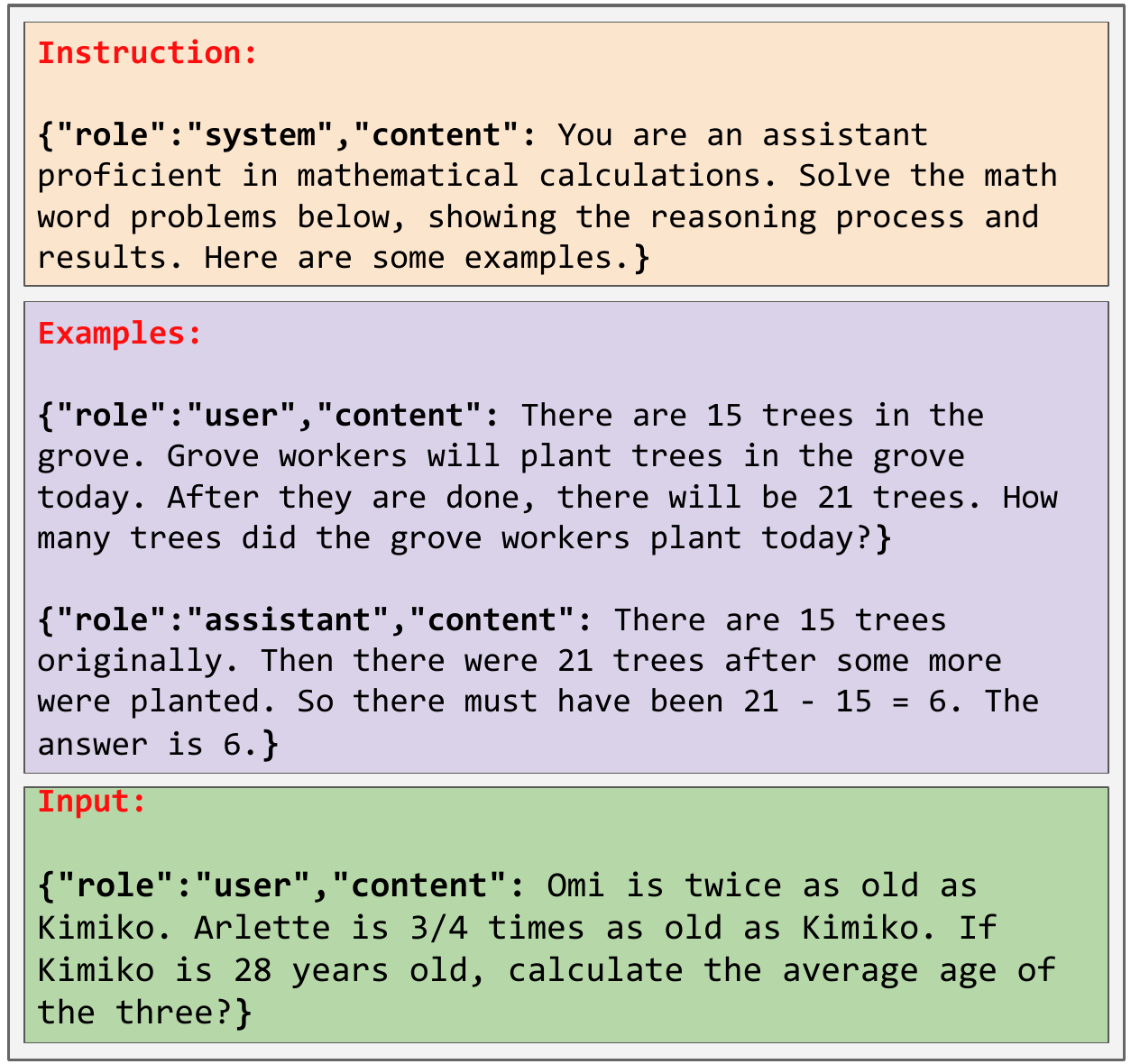}
    \caption{Illustration of the initial prompt for GSM8K with GPT3 \textbf{(left)} and GPT3.5 \textbf{(right)} as backbone model}
    \label{fig:initial-prompt-gpt}   
\end{figure}

\paragraph{Experimental Details}
\label{appendix:prompts-details}
The initial few-shot-CoT prompts provided followed the \textbf{Instruction + Examples} format, consisting of a task-agnostic instruction and several positive examples (details are shown in Figure \ref{fig:initial-prompt-gpt}, which is consistent with the setting described in \citep{prasad2022grips}.
The \textbf{Examples} part of the initial prompts is the same as the few-shot exemplars by \citet{wei2022cot}. The \textbf{Instruction} provides a general description of the downstream task, although its specific content may vary depending on the backbone models used.
The new prompt candidates are generated by applying edits to both the instruction and examples with equal probability during the searching process, i.e $Prob\{\text{edit the instruction}\} = Prob\{\text{edit the examples}\} = 1/2$.

For GPT3, the \textbf{Instruction} part of the prompt follows the zero-shot-CoT approach proposed by \citet{kojima2022zerocot}. Meanwhile, for GPT3.5, the \textbf{Instruction} part of the initial prompts is manually crafted by referring to the API reference documentation available on the GPT3.5 in the OpenAI Platform\footnote{https://platform.openai.com/docs/guides/gpt/chat-completions-api}.

\begin{figure}[h]
    \centering
    \includegraphics[scale=0.25]{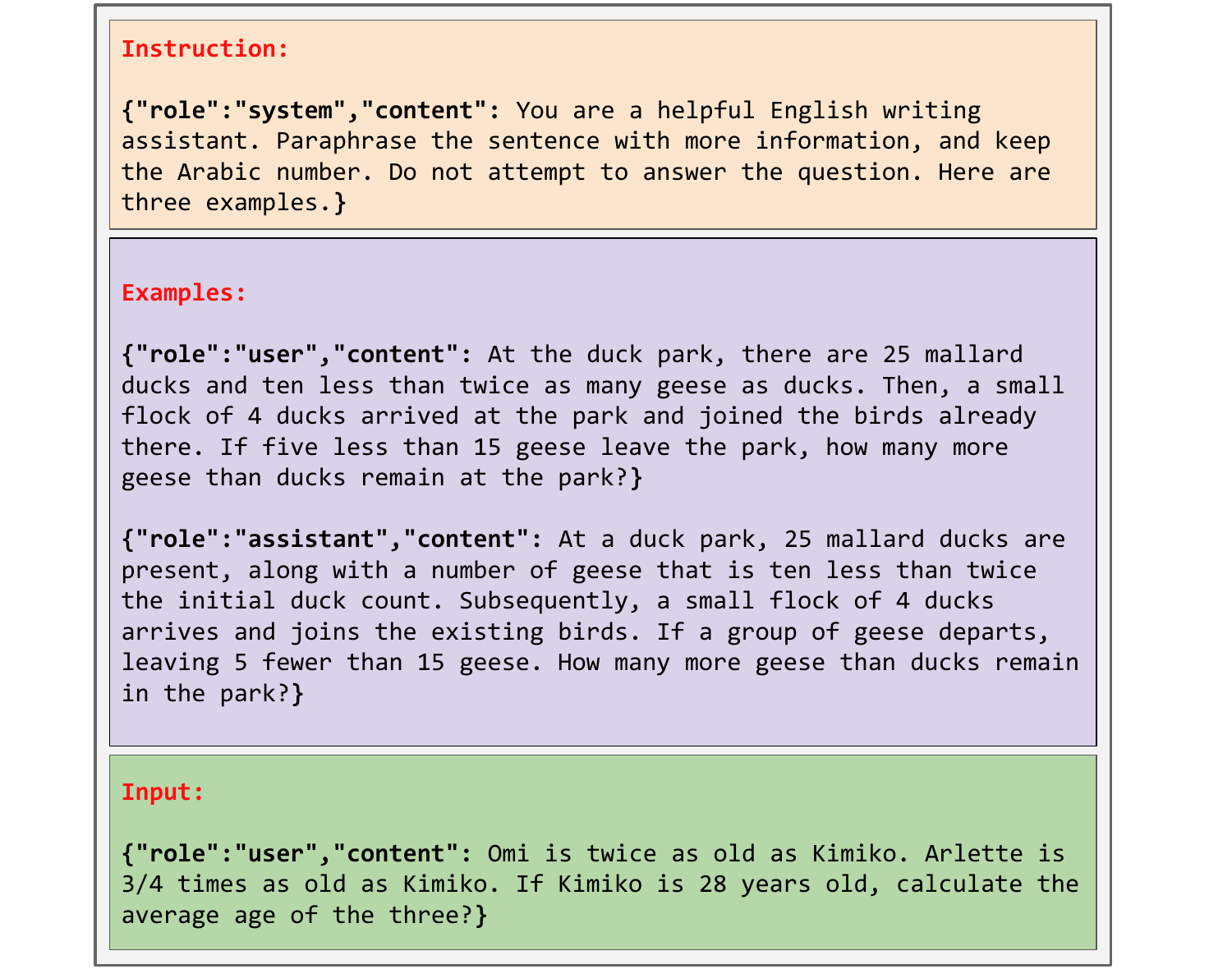}
    \includegraphics[scale=0.25]{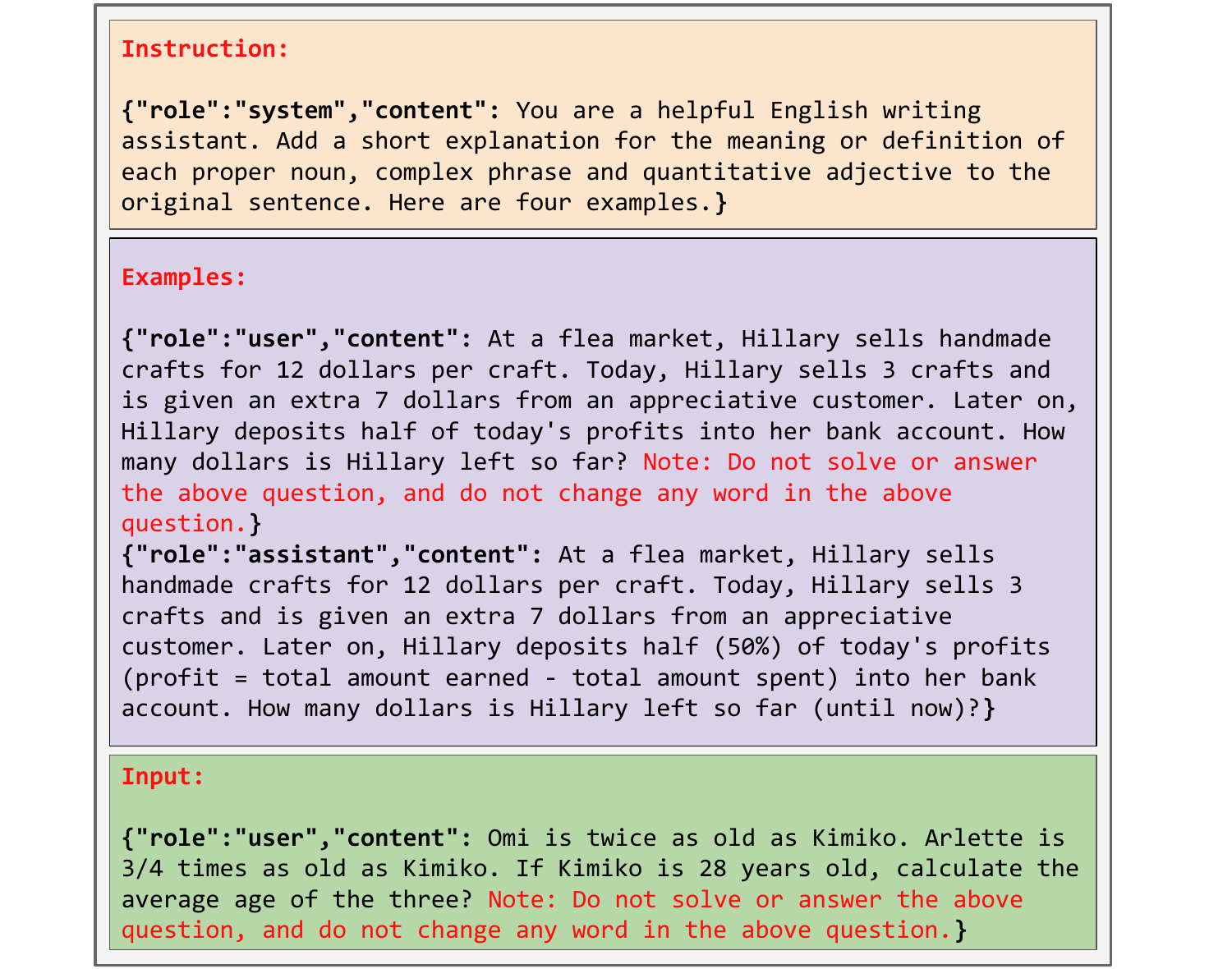}
    \caption{Illustration of the improved \texttt{add} operation and revision with GPT3.5 as backbone model}
    \label{fig:add-revision-prompt-gpt3.5}   
\end{figure}

Moreover, Figure \ref{fig:add-revision-prompt-gpt3.5} illustrates the template for the prompt used of the improved \texttt{add} operation and revisions, leveraging the GPT3.5 as the underlying model.

\paragraph{Experimental Cost}

While producing few-shot-CoT prompts, we do not consider the API call limit; however, we still take into account the estimated cost of the experiments. The costs for GPT3 (\textit{text-curie-001}) and GPT3.5 (\textit{gpt-3.5-turbo-0301}) are nearly identical as they share the same pricing structure (\$0.002 per 1k tokens). On average, each task typically involves a cost of approximately \$25 and concludes after approximately 12 hours of execution when searching for few-shot-CoT prompts with a single thread.

\section{Algorithm Details}
\label{appendix:alg-details}

To help users better understand and utilize the package, here we present the main idea and pseudo-codes of all included algorithms so far.

\subsection{Hill Climbing}

Hill climbing is a greedy algorithm that aims to find the local optimum of a given objective, e.g. the score function. Each time the algorithm explores a fraction of possible prompt changes and selects one that improves the score.

In our version, we select the one with the most improvements to further speed up the optimization process, as shown in Algorithm~\ref{alg:GrIPS}. Notice that in the paradigm of EvoPrompt, GrIPS can also be viewed as a special version of Hill Climbing, given it greedily improves the prompt via edit operations.

\begin{algorithm}[h!]
\begin{algorithmic}[1]
\caption{GrIPS / Hill Climbing}
\State $base \gets init$ \Comment{Initialize base candidate}
\State $s_{base} \gets \texttt{score}(base)$ \Comment{Score using examples in $\mathcal{S}$}
\State $\Omega \gets \{ \texttt{del}, \texttt{swap}, \texttt{par}, \texttt{add} \}$
\Comment{Set of edit operations}
\State $\rho \gets P$ \Comment{Patience for early-stop}
\For{$i = 1, \cdots, n$} \Comment{$n$: number of iterations}
  \For{$j = 1, \cdots, m$} \Comment{$m$: number of candidates}
    \State Sample $e_1, \cdots, e_l \in \Omega$
    \Comment{$l$ edits per candidate}
    \State $\mathcal{C}[j] \gets \texttt{edit}(base, e_1 \circ \cdots \circ e_l)$ 
    \State $s[j] \gets \texttt{score}(\mathcal{C}[j])$
    \Comment{Score above candidate}
  \EndFor
  \State $k \gets \text{argmax}_j s[j]$
  \State $best \gets \mathcal{C}[k]$ \Comment{Chosen Candidate}
  \State $s_{best} \gets s[k]$ \Comment{Score of chosen candidate}
  \If{$s_{best} > s_{base}$} 
    \Comment{Candidate better than base}
    \State $base \gets best$
    \Comment{Use this candidate in the next step}
    \State $s_{base} \gets s_{best}$
    \Comment{Update base score}
    \State $\rho \gets P$
    \Comment{Refresh patience}
  \Else
    \If{$\rho > 0$} \Comment{Patience not exhausted}
      \State \textbf{decrement} $\rho$
      \State \textbf{continue}
      \Comment{Continue search with same base}
    \Else
      \State \Return base
      \Comment{Early-stop criteria met}
    \EndIf
  \EndIf
\EndFor
\State \Return base
\Comment{Search terminates after the last iteration}
\label{alg:GrIPS}
\end{algorithmic}
\end{algorithm}

\subsection{Simulated Annealing}

Different from hill climbing, simulated annealing~\citep{kirkpatrick1983optimization} allows a small probability of deviation when the prompt change does not yield instant improvement, thus capable of escaping from local optimums and finding better solutions. The probability of deviation is controlled by a hyperparameter called ``temperature'' $T(i)$ at iteration $i$, where the lower temperature is, smaller the deviation probability is, less random the algorithm behavior becomes. When the temperature $T(i) = 0$, simulated annealing becomes deterministic and degenerates to hill climbing.

As shown in Algorithm~\ref{alg:simulated-annealing}, it only differs from hill climbing in one line, where it utilizes the provided temperature $T(i)$ to decide the deviation probability. We use exponential decay $T(i) = 10 \exp(-i / 5)$ as our default temperature schedule in the package.

\begin{algorithm}[t]
\begin{algorithmic}[1]
\caption{Simulated Annealing}
\State $base \gets init$ \Comment{Initialize base candidate}
\State $s_{base} \gets \texttt{score}(base)$ \Comment{Score using examples in $\mathcal{S}$}
\State $\Omega \gets \{ \texttt{del}, \texttt{swap}, \texttt{par}, \texttt{add} \}$
\Comment{Set of edit operations}
\State $\rho \gets P$ \Comment{Patience for early-stop}
\For{$i = 1, \cdots, n$} \Comment{$n$: number of iterations}
  \For{$j = 1, \cdots, m$} \Comment{$m$: number of candidates}
    \State Sample $e_1, \cdots, e_l \in \Omega$
    \Comment{$l$ edits per candidate}
    \State $\mathcal{C}[j] \gets \texttt{edit}(base, e_1 \circ \cdots \circ e_l)$ 
    \State $s[j] \gets \texttt{score}(\mathcal{C}[j])$
    \Comment{Score above candidate}
  \EndFor
  \State $k \gets \text{argmax}_j s[j]$
  \State $best \gets \mathcal{C}[k]$ \Comment{Chosen Candidate}
  \State $s_{best} \gets s[k]$ \Comment{Score of chosen candidate}
  \If{$s_{best} > s_{base}$ \textbf{or} $\exp(\frac{s_{best} - s_{base}}{T(i)}) \ge \texttt{Random}(\mathcal{U}(0,1))$} 
    \Comment{Candidate better than base}
    \State $base \gets best$
    \Comment{Use this candidate in the next step}
    \State $s_{base} \gets s_{best}$
    \Comment{Update base score}
    \State $\rho \gets P$
    \Comment{Refresh patience}
  \Else
    \If{$\rho > 0$} \Comment{Patience not exhausted}
      \State \textbf{decrement} $\rho$
      \State \textbf{continue}
      \Comment{Continue search with same base}
    \Else
      \State \Return base
      \Comment{Early-stop criteria met}
    \EndIf
  \EndIf
\EndFor
\State \Return base
\Comment{Search terminates after the last iteration}
\label{alg:simulated-annealing}
\end{algorithmic}
\end{algorithm}

\subsection{Genetic Algorithm (Mutation Only)}

Genetic algorithm~\citep{holland1992genetic} was inspired by biological evolution during the nature selection process, where each prompt is treated as a ``gene'' and the score function serves as the criterion of nature selection.

In a simplified version of genetic algorithm, as illustrated in Algorithm~\ref{alg:ga-m}, only mutation operations are adopted to change the prompt. A population of prompts $\bW$ are maintained during the optimization process, indicating all the possible exploring directions of the algorithm. At each iteration, we select a high-quality prompt from all those exploring directions via tournament selection. Intuitively, the more tournament round $k$ we adopt, the higher quality of the selected prompt, but the less explorative the algorithm behavior becomes. The selected prompt is then mutated to generate a new prompt.

As the algorithm proceeds, the average quality of prompts in the population is expected to improve. In fact, given sufficient number of iterations, the algorithm is capable of finding the global optimum.

\begin{algorithm}[t]
\begin{algorithmic}[1]
\caption{Genetic Algorithm (mutation only)}
\label{alg:ga-m}
\State $base \gets init$ \Comment{Initialize base candidate}
\State $s_{base} \gets \texttt{score}(base)$ \Comment{Score using examples in $\mathcal{S}$}
\State $\Omega \gets \{ \texttt{del}, \texttt{swap}, \texttt{par}, \texttt{add} \}$
\Comment{Set of edit operations}
\State $\rho \gets P$ \Comment{Patience for early-stop}
\State \tblack{$\bW \gets \{ \langle base, s_{base} \rangle \}$}
\State \tblack{$result \gets base$}
\Comment{Initialize population}
\For{$i = 1, 2, \cdots, n$} \Comment{$n$: number of iterations}
  \State \tblack{$\bS \gets \emptyset$}
  \Comment{Examples that are going to battle in the tournament}
  \For{\tblack{$j = 1, 2, \cdots, k$}} \Comment{$k$: tournament selection hyperparameter}
    \State \tblack{$\langle parent, s_{parent} \rangle \gets \texttt{Random}(\bW)$}
    \State \tblack{$\bS \gets \bS + \{ \langle parent, s_{parent} \rangle \} $}
  \EndFor
  \State \tblack{$base \gets \text{argmax}_{\{ parent | \langle parent, s_{parent} \rangle \in \bS\} } s_{parent}$}
  \For{$j = 1, \cdots, m$} \Comment{$m$: number of candidates}
    \State Sample $e_1, \cdots, e_l \in \Omega$
    \Comment{$l$ edits per candidate}
    \State $\mathcal{C}[j] \gets \texttt{edit}(base, e_1 \circ \cdots \circ e_l)$ 
    \State $s[j] \gets \texttt{score}(\mathcal{C}[j])$
    \Comment{Score above candidate}
  \EndFor
  \State $k \gets \text{argmax}_j s[j]$
  \State $best \gets \mathcal{C}[k]$ \Comment{Chosen Candidate}
  \State $s_{best} \gets s[k]$ \Comment{Score of chosen candidate}
  \State \tblack{$\bW \gets \bW + \{ \langle best, s_{best} \rangle \}$}
  \If{\tblack{$s_{best} > s_{result}$}}
    \Comment{Candidate better than best result so far}
    \State \tblack{$result \gets best$}
    \State \tblack{$s_{result} \gets s_{best}$}
    \State $\rho \gets P$
    \Comment{Refresh patience}
  \Else
    \If{$\rho > 0$} \Comment{Patience not exhausted}
      \State \textbf{decrement} $\rho$
      \State \textbf{continue}
      \Comment{Continue search with same base}
    \Else
      \State \Return \tblack{$result$}
      \Comment{Early-stop criteria met}
    \EndIf
  \EndIf
\EndFor
\State \Return \tblack{$result$}
\Comment{Search terminates after the last iteration}
\end{algorithmic}
\end{algorithm}

\subsection{Genetic Algorithm (with Crossover)}

The general version of genetic algorithm~\citep{holland1992genetic} is realized in Algorithm~\ref{alg:ga-c}. Besides the mutation operation, a different operation called ``crossover'' is introduced, where we allow new prompts to be generated from two ``parent prompts''. The new prompts will inherit partial segments from both parent prompts and be further changed with mutation operations.

Intuitively speaking, crossover operation represents ``global'' changes in the search space, while mutation operation represents ``local'' changes, hence it is to be expected that the genetic algorithm with crossover is less efficient than its simplified counterpart, but more explorative and capable of finding better solutions with sufficient iterations.

\begin{algorithm}[t!]
\footnotesize
\textbf{Hyperparameters:} Number of iterations $n$, population size $N_p$, number of offsprings each generation $N_{\text{offspring}}$, mutation probability $p_{mutation}$.
\begin{algorithmic}[1]
\caption{Genetic Algorithm (with crossover)}
\label{alg:ga-c}
\State $base \gets init$ \Comment{Initialize base candidate}
\State $s_{base} \gets \texttt{score}(base)$ \Comment{Score using examples in $\mathcal{S}$}
\State $\Omega \gets \{ \texttt{del}, \texttt{swap}, \texttt{par}, \texttt{add} \}$
\Comment{Set of edit operations}
\State $\rho \gets P$ \Comment{Patience for early-stop}
\State \tblack{$\bW_0 \gets \{ \langle base, s_{base} \rangle \times N_p \}  $}
\Comment{$N_p$ copies of the initial prompt}
\State \tblack{$result \gets base$}
\For{$i = 1, 2, \cdots, n$} \Comment{$n$: number of iterations}
  \State \tblack{$\bW_i \gets \bW_{i-1}$}
  \For{\tblack{$j = 1, 2, \cdots, N_{\text{offspring}}$}}
  \Comment{===== Crossover}
    \State \tblack{$\langle parent_1, s_{parent_1} \rangle \gets$ \Call{binary\_tournament\_selection}{$\bW_{i-1}$}}
    \State \tblack{$\langle parent_2, s_{parent_2} \rangle \gets$ \Call{binary\_tournament\_selection}{$\bW_{i-1}$}}
    \State \tblack{$\text{offspring} \gets$ \Call{crossover}{$parent_1, parent_2$}}
    \If{\tblack{$\text{offspring} \not \in \bW_i$}}
      \State \tblack{$s_{\text{offspring}} \gets \texttt{score}(\text{offspring})$}
      \State \tblack{$\bW_i \gets \bW_i + \{ \langle \text{offspring}, s_{\text{offspring}} \rangle \}$}
    \EndIf
  \EndFor
  \State \tblack{$\bW_i \gets \texttt{top-}N_p(\bW_i)$}
  \Comment{Removes prompts with bad scores, reserves only best $N_p$ prompts}
  \State \tblack{$\rho, result, flag_{stop} \gets$ \Call{update\_result}{$\rho, result, \bW_i$}}
  \If {\tblack{$flag_{stop}$}}
    \Return \tblack{$result$}
  \EndIf
  \\
  \State $\tblack{\bW'_i \gets \bW_i}$
  \For{\tblack{$\forall \langle base, s_{base} \rangle \in \bW_i$}}
  \Comment{===== Mutation}
    \If{\tblack{$p_{mutation} \ge \texttt{Random}([0, 1])$}}
      \State Sample $e_1, \cdots, e_l \in \Omega$
      \Comment{$l$ edits per candidate}
      \State $\tblack{mutated} \gets \texttt{edit}(base, e_1 \circ \cdots \circ e_l)$ 
      \State $\tblack{s_{mutated} \gets \texttt{score}(mutated)}$
      \Comment{Score above candidate}
      \State \tblack{$\bW'_i \gets \bW'_i + \{ \langle mutated, s_{mutated} \rangle \}$}
    \EndIf
  \EndFor
  \State \tblack{$\bW_i \gets \texttt{top-}N_p(\bW'_i)$}
  \Comment{Removes prompts with bad scores, reserves only best $N_p$ prompts}
  \State \tblack{$\rho, result, flag_{stop} \gets$ \Call{update\_result}{$\rho, result, \bW_i$}}
  \If {\tblack{$flag_{stop}$}}
    \Return \tblack{$result$}
  \EndIf
\EndFor
\State \Return \tblack{$result$}
\Comment{Search terminates after last iteration}
\\
\Function{binary\_tournament\_selection}{$\bW$}
  \State \tblack{$\bS \gets \emptyset$}
  \For{\tblack{$i = 1, 2$}}
    \State \tblack{$\langle parent, s_{parent} \rangle \gets \texttt{Random}(\bW)$}
    \State \tblack{$\bS \gets \bS + \{ \langle parent, s_{parent} \rangle \} $}
  \EndFor
  \State \tblack{\Return $\text{argmax}_{\{ parent | \langle parent, s_{parent} \rangle \in \bS\} } s_{parent}$}
\EndFunction
\\
\Function{crossover}{$parent_1, parent_2$}
  \State \tblack{$[\bt_{1,1}, \bt_{1,2}, \dots, \bt_{1,L_1}] \gets parent_1$}
  \State \tblack{$[\bt_{2,1}, \bt_{2,2}, \dots, \bt_{2,L_2}] \gets parent_2$}
  \State \tblack{$split \gets \texttt{Random}(\{0, 1, 2, \dots, \max(L_1, L_2)\})$}
  \State $\tblack{\text{offspring} \gets [\bt_{1,1}, \dots, \bt_{1, split}, \bt_{2,split+1}, \dots, \bt_{2,L_2}}]$
  \State \Return \tblack{\text{offspring}}
\EndFunction
\\
\Function{update\_result}{$\rho, result, \bW$}
  \State \tblack{$flag_{stop} \gets \textbf{false}$}
  \State \tblack{$best \gets \text{argmax}_{\{ prompt | \langle prompt, s_{prompt} \rangle \in \bW\} } s_{prompt}$}
  \If{\tblack{$s_{best} > s_{result}$}}
    \Comment{Candidate better than best result so far}
    \State \tblack{$result \gets best$}
    \State \tblack{$s_{result} \gets s_{best}$}
    \State $\rho \gets P$
    \Comment{Refresh patience}
  \Else
    \If{$\rho > 0$} \Comment{Patience not exhausted}
      \State \textbf{decrement} $\rho$
    \Else
      \State \tblack{$flag_{stop} \gets \textbf{true}$}
      \Comment{Early-stop criteria met}
    \EndIf
  \EndIf
  \State \Return $\rho, result, flag_{stop}$
\EndFunction
\end{algorithmic}
\end{algorithm}

\subsection{Tabu Search}

Greedy search algorithms tend to be stuck in a local optimum or near a local optimum. Tabu Search~\citep{glover1986tabu} effectively compensates for this shortcoming by introducing the Tabu list: any candidate that was visited recently is forbidden to visit again. This allows the optimization process to leave the local ``tabu'' area and explore more.

In Algorithm~\ref{alg:tabu-search} we adopt a simple Tabu rule of direct comparison: with a high probability, a generated prompt that exactly matches any prompts in the Tabu list will be treated as invalid and discarded, where the Tabu list only retains the $N_{tabu}$ latest generated prompts.

\begin{algorithm}[htbp!]
\begin{algorithmic}[1]
\caption{Tabu Search}
\label{alg:tabu-search}
\State $base \gets init$ \Comment{Initialize base candidate}
\State $s_{base} \gets \texttt{score}(base)$ \Comment{Score using examples in $\mathcal{S}$}
\State $\Omega \gets \{ \texttt{del}, \texttt{swap}, \texttt{par}, \texttt{add} \}$
\Comment{Set of edit operations}
\State $\rho \gets P$ \Comment{Patience for early-stop}
\State {$result \gets base$}
\Comment{Initialize population}
\State $w_0 \gets base$
\State $\mathcal{T} \gets \{ w_0 \}$ \Comment{Initialize $\mathcal{T}$}
\For{$i = 1, 2, \cdots, K$} \Comment{$n$: number of iterations}

    \For{$j = 1, 2, \cdots, n$} \Comment{$k$: number of the new candidate}
        \State Sample $e_1, \cdots, e_l \in \Omega$
        \Comment{$l$ edits per candidate}
        \State $\mathcal{C}[j] \gets \texttt{edit}(base, e_1 \circ \cdots \circ e_l)$ 
        \State $s[j] \gets \texttt{score}(\mathcal{C}[j])$
        \Comment{Score above candidate}
    \EndFor
    \State $\mathcal{W}'_j \gets$ \Call{Tabu}{$\mathcal{T},s[j], TEMP$} = 0
    \State $k \gets \text{argmax}_{l \in \mathcal{W}'_j} s[j]$
    \State $best \gets \mathcal{C}[k]$ \Comment{Chosen Candidate}
    \State $s_{best} \gets s[k]$ \Comment{Score of chosen candidate}
    \State{$\mathcal{T} \gets \mathcal{T} + s[k]$} 
    \Comment{Update $\mathcal{T}$}
     \If{$s_{best} > s_{result}$}
        \Comment{Candidate better than best result so far}
        \State $result \gets best$
        \State $s_{result} \gets s_{best}$
        \State $\rho \gets P$
        \Comment{Refresh patience}
      \Else
        \If{$\rho > 0$} \Comment{Patience not exhausted}
          \State \textbf{decrement} $\rho$
          \State \textbf{continue}
          \Comment{Continue search with same base}
        \Else
          \State \Return $result$
          \Comment{Early-stop criteria met}
        \EndIf
      \EndIf
    \If{$|\mathcal{T}| > N_{tabu}$}
        \State $\mathcal{T} \gets \mathcal{T} - w_{k-N_{tabu}}$\Comment{Keep Short-term tabus}
    \EndIf
\EndFor
\State \Return $result$
\Comment{Search terminates after last iteration}

\Function{Tabu}{$\mathcal{T}$,$w$, $TEMP$}
\If{$w \in \mathcal{T}$}
    \If{$TEMP \ge \texttt{Random}(\cU(0, 1))$}
        \State \Return 0
    \Else \State \Return 1
    \EndIf
\Else
\State \Return 0
\EndIf
\EndFunction
\end{algorithmic}
\end{algorithm}

\subsection{Harmony Search}

Harmony Search~\citep{geem2001harmonysearch} mimics the improvision procedure of musicians, where a new prompt is generated via combining segments of all stored prompts in the harmony memory. A pitch finetuning process is then conducted, allowing further refinement of the prompt. The full details are available in Algorithm~\ref{alg:harmony-search}.

\begin{algorithm}[htbp!]
\begin{algorithmic}[1]
\caption{Harmony Search}
\label{alg:harmony-search}
\State $base \gets init$ \Comment{Initialize base candidate}
\State $s_{base} \gets \texttt{score}(base)$ \Comment{Score using examples in $\mathcal{S}$}
\State $\Omega \gets \{ \texttt{del}, \texttt{swap}, \texttt{par}, \texttt{add} \}$
\State $\Omega_{small} \gets \{ \texttt{par} \}$
\Comment{Set of edit operations}
\State $\rho \gets P$ \Comment{Patience for early-stop}
\State $\bW_0 \gets \{ \langle base, s_{base} \rangle  \} $
\State $result \gets base$
\Comment{Initialize population}

\For{$i = 1, 2, \cdots, n$} \Comment{$n$: number of iterations}
  \State $\bW' \gets \emptyset$
  \Comment{Candidates generated at this iteration}
  \For{$c = 1, 2, \cdots, k$}
    \State $\bw' \gets$ \Call{generate\_candidate}{$\bW_{i-1}, \Omega_{small}, \Omega$}
    \State $s_{\bw'} \gets \texttt{score}(\bw')$
    \Comment{Score the new candidate}
    \State $\bW' \gets \bW' + \{ \langle \bw', s_{\bw'} \rangle \}$
  \EndFor
  \State $\rho, result, flag_{stop} \gets$ \Call{update\_result}{$\rho, result, \bW'$}
  \If {$flag_{stop}$}
    \Return $result$
  \EndIf
  \State $\bW_i \gets \texttt{top-}N_H(\bW_{i-1} \cup \bW')$
  \Comment{Removes prompts with bad scores, reserves only best $N_H$ prompts}
\EndFor
\State \Return $result$
\Comment{Search terminates after last iteration}
\Function{generate\_candidate}{$\bW, \Omega_{small}, \Omega$}
    \State $\bw' \gets [ \ ]$
    \Comment{Empty string/list}
    \For{$j = 1, 2, \cdots, k_s$} \Comment{$k_s$: number of segments (pitches)}
      \State $\langle \bw, * \rangle \gets \texttt{Random}(\bW)$
      \Comment{Randomly sample an existing prompt for $j$-th segment}
      \State $L \gets |\bw|$
      \Comment{Length of the reference prompt}
      \State $[w_0, w_1, \dots, w_{L-1}] \gets \bw$
      \State $start \gets \lceil\frac{j-1}{k_s} \cdot L\rceil$
      \State $end \gets \lceil\frac{j}{k_s} \cdot L\rceil - 1$
      \State $\bw_{segment} \gets [w_{start}, w_{start+1}, \dots, w_{end}]$
      \If{$HMCR \ge \texttt{Random}(\cU(0, 1))$}
        \If{$PAR \ge \texttt{Random}(\cU(0, 1))$}
        \Comment{A little different random segment (pitch)}
          \State Sample $e_1, \cdots, e_l \in \Omega_{small}$
          \Comment{Only replace phrases with their synonyms}
          \State $\bw_{segment}\gets $  $\texttt{edit}(\bw_{segment}, e_1 \circ \cdots \circ e_l)$
        \EndIf
      \Else
      \Comment{A largely different random segment (pitch)}
        \State Sample $e_1, \cdots, e_l \in \Omega$
        \Comment{A big change in the original segment}
        \State $\bw_{segment} \gets  \texttt{edit}(\bw_{segment}, e_1 \circ \cdots \circ e_l)$
      \EndIf
      \State $\bw' = \bw' +  \bw_{segment}$
      \Comment{Here ``$+$'' means concatenation.}
    \EndFor
    \State \Return $\bw'$
\EndFunction
\Function{update\_result}{$\rho, result, \bW$}
  \State $flag_{stop} \gets \textbf{false}$
  \State $best \gets \text{argmax}_{\{ prompt | \langle prompt, s_{prompt} \rangle \in \bW\} } s_{prompt}$
  \If{$s_{best} > s_{result}$}
    \Comment{Candidate better than best result so far}
    \State $result \gets best$
    \State $s_{result} \gets s_{best}$
    \State $\rho \gets P$
    \Comment{Refresh patience}
  \Else
    \If{$\rho > 0$} \Comment{Patience not exhausted}
      \State \textbf{decrement} $\rho$
    \Else
      \State $flag_{stop} \gets \textbf{true}$
      \Comment{Early-stop criteria met}
    \EndIf
  \EndIf
  \State \Return $\rho, result, flag_{stop}$
\EndFunction
\end{algorithmic}
\end{algorithm}

\section{Hyperparameter Setting}
For all the experiments, the number of edits that performs on the candidate prompts ($num\_compose$) and the number of the newly generated candidates ($num\_candidate$) in each iteration follows $num\_compose \in \{1,2\}$ and $num\_candidate \in \{5, 10\}$. The population sizes, and tournament selections $k$ of \plumgam{}, \plumgac{} and \plumgam{}* are all set to $10$ and $3$ respectively. Meanwhile, the mutation rate $p_{\text{mutation}}$ is set to $0.5$ for \plumgac{}. The $N_{\text{tabu}}$ is configured to $5$ for \plumts{}. Additionally, the configuration for the harmony search memory ($N_H$), the number of segments ($k_s$), harmony memory considering rate ($HMCR$), and pitching adjust rate ($PAR$) of \plumhs{}  and \plumhs{}* is as follows: $10$, $5$, $0.4$, and $0.5$, respectively. Furthermore, the temperatures of GPT-3 are configured to $0$ in our experiments.

In the context of general prompt learning, disregarding the API call limit allows us to set the maximum iteration and patience parameters to $50$ and $7$ respectively, which ensures that \plumgam{} and \plumgac{} have a sufficient number of iterations to generate relatively optimal solutions.

When it comes to few-shot-CoT prompt learning, setting a very large value for the maximum iterations is not recommended due to the potential cost implications, which is primarily because the number of tokens generated in each iteration tends to be substantial. And for \plumgam{}* and \plumhs{}*, effective results can be obtained with maximum iterations ranging from $5$ to $15$, coupled with a patience level ranging from $2$ to $5$. For the experimental results shown in Table \ref{tab:few-shot-cot-improv}, we configured the maximum iterations to be $10$ and deactivated the patience parameter.

For text-to-image generation tasks, to accommodate a wide range of application scenarios, the initial prompts for image generation are kept short (merely $5$-$15$ words). Given that stable diffusion requires approximately $15$ seconds (on a NVIDIA A40 GPU) to produce each image, and the scoring process necessitates the generation of two images for direct comparison, it has been observed that after $6$-$7$ iterations, most prompts reach a point of convergence where the best score is achieved. Consequently, it is advisable to limit the number of iterations to $5$-$15$ and set the patience parameter between $2$-$5$. In the experimental results presented in Table \ref{tab:extra_plum_stable_diffusion}, we set the maximum iterations to $10$ and deactivated the patience parameter. Furthermore, considering the prompt's brevity, the initial $k_s$ value was set to $2$ in the experiments to guarantee that each segment targeted for modification includes at least one tag. Additionally, the Harmony Memory Considering Rate ($HMCR$), Pitch Adjustment Rate ($PAR$), and  Harmony Search Memory ($n_H$) were set to $0.4$, $0.5$, and $10$ respectively.

\paragraph{Sensitiveness of \plumhs{} Hyperparameters}

\begin{figure}
    \centering
    \begin{subfigure}{0.4\textwidth}
      \includegraphics[width=1.1\linewidth]{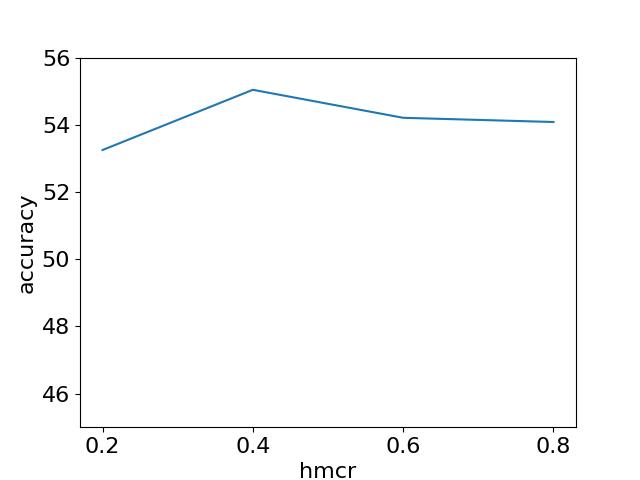}
      \caption{$HMCR$}
      \label{fig:abl-hmcr}
    \end{subfigure}%
    \begin{subfigure}{0.4\textwidth}
      \includegraphics[width=1.1\linewidth]{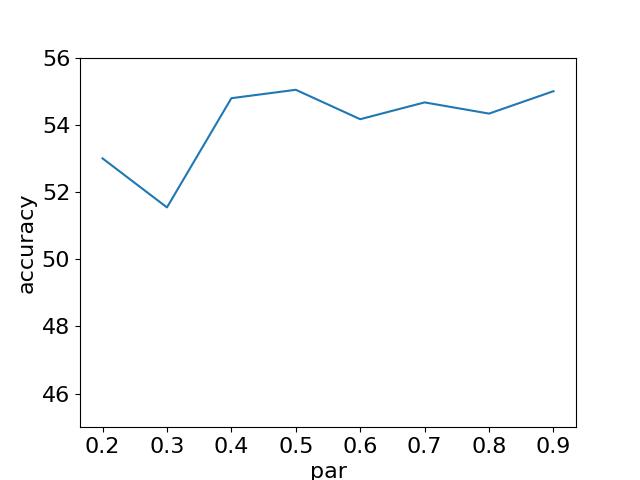}
      \caption{$PAR$}
      \label{fig:abl-par}
    \end{subfigure}%
    \\
    \begin{subfigure}{0.4\textwidth}
      \includegraphics[width=1.1\linewidth]{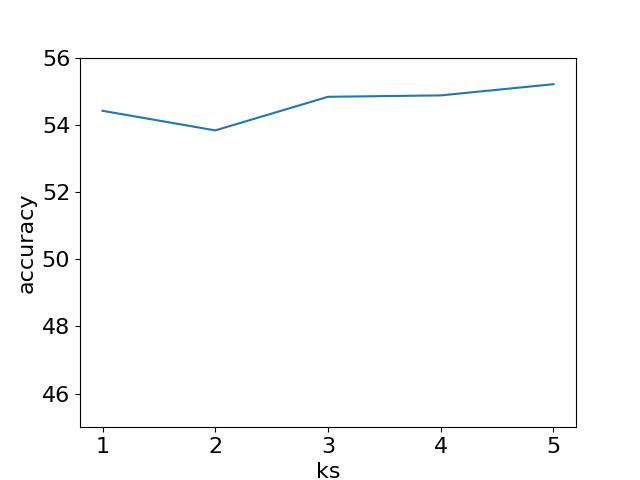}
      \caption{$k_s$}
      \label{fig:abl-ks}
    \end{subfigure}
    \begin{subfigure}{0.4\textwidth}
      \includegraphics[width=1.1\linewidth]{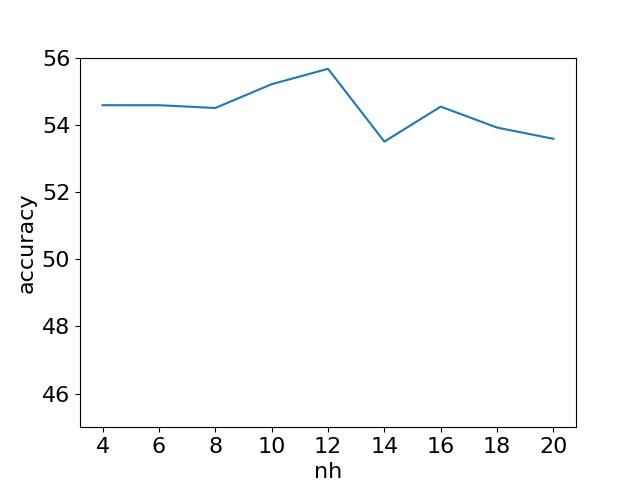}
      \caption{$N_H$}
      \label{fig:abl-nh}
    \end{subfigure}
    \caption{The effects of the parameters of \plumhs{} with GPT2-large as backbone.}
    \label{fig:abl-study}
\end{figure}

To gain deeper insights into the impact of key hyperparameters in Harmony Search and guide their effective tuning, we conducted an ablation study on $HMCR$, $k_s$ and $N_H$ for \plumhs{} with GPT2-large as the backbone model. As shown in Figure~\ref{fig:abl-study}, both the Harmony Memory Consideration Rate $HMCR$ and the Harmony Memory size $N_H$ achieve optimal performance with a moderate value. This trend aligns with the core principles of Harmony Search (Algorithm \ref{alg:harmony-search}). Low $HMCR$ encourages the algorithm perform random modifications, which results in introducing too much difference. Conversely, high $HMCR$ over-relies on past prompts in the harmony memory, which lacks diversity. As for Harmony Memory size $N_H$, it is similar to the population size in genetic algorithms. Small $N_H$ restricts exploration such that the algorithm behaves similarly to greedy search and lacks diversity. While, large $N_H$ deteriorates the quality of memorized prompts, and slows down the search efficiency. Notably, the performance exhibits stable and high performance when $PAR$ is set to $0.4$ or above.

For $k_s$, we find that a larger value introduces superior performance. This is because larger $k_s$ leads to finer granularity of segments, which allows a finer level of tuning on prompts. In this paper, the upper bound of $k_s$ is considered as $5$ since the shortest sample in our datasets only contains $5$ segments of words.

\section{Licenses}

For general prompt learning tasks, the dataset Natural-Instructions v2.6~\citep{mishra2022cross} is released under Apache-2.0 license. While for Chain-of-Thought prompt learning, the datasets ASDiv~\citep{miao-etal-2020-diverse}, SVAMP~\citep{Patel2021Mar},
CSQA~\citep{talmor-etal-2019-commonsenseqa} are released under CC-BY-NC 4.0, MIT, and CC BY-SA 4.0 licenses respectively.

\end{document}